\pdfoutput=1

\documentclass[11pt]{article}

\usepackage[preprint]{acl}

\usepackage{times}
\usepackage{latexsym}
\usepackage{amsmath}
\usepackage{amsthm,amsmath,amssymb}
\usepackage{mathrsfs}
\usepackage{graphicx}
\usepackage{tabularx}
\usepackage{color}
\usepackage{booktabs}
\usepackage{tablefootnote}
\usepackage{pifont} 
\usepackage{arydshln} 
\usepackage{multirow}
\usepackage{hhline}
\usepackage[T1]{fontenc}
\DeclareUnicodeCharacter{0307}{u}

\usepackage[utf8]{inputenc}

\usepackage{microtype}

\usepackage{inconsolata}
\usepackage{tcolorbox}
\definecolor{hidden-draw}{RGB}{20,68,106}
\definecolor{hidden-pink}{RGB}{255,245,247}
\definecolor{maroon}{RGB}{148,78,99}
\definecolor{hidden-white}{RGB}{245,238,230}
\definecolor{hidden-yellow}{RGB}{255,248,227}
\definecolor{hidden-orange}{RGB}{243,215,202}
\definecolor{xm-purple}{RGB}{216, 218, 237}
\definecolor{xm-grey}{RGB}{242,242,242}
\newtcolorbox[list inside=prompt,auto counter]{prompt}[1][]{
    colbacktitle=xm-purple!90,
    colback =xm-grey!30,
    coltitle=black,
    fontupper=\footnotesize,
    boxsep=5pt,
    left=0pt,
    right=0pt,
    top=0pt,
    bottom=0pt,
    boxrule=0.5pt,
    #1,
}
\title{PROM: A Phrase-level Copying Mechanism with Pre-training for Abstractive Summarization}

\author{
Xinbei Ma\textsuperscript{\rm 1,2,${\ast}$ \thanks{${\ast}$Work done during an internship at $^3$Microsoft Research Asia. \# Equal contribution. \dag Corresponding author. }},
Yeyun Gong\textsuperscript{\rm 3, \#, \dag},
Pengcheng He\textsuperscript{\rm \#},
Hai Zhao\textsuperscript{\rm 1,2,\dag \thanks{This paper was partially supported by Joint Research Project of Yangtze River Delta Science and Technology Innovation Community (No. 2022CSJGG1400).}},
Nan Duan\textsuperscript{\rm 3}\\
  $^1$Department of Computer Science and Engineering,  Shanghai Jiao Tong University
  \\ $^2$Key Laboratory of Shanghai Education Commission for Intelligent Interaction \\
and Cognitive Engineering, Shanghai Jiao Tong University \\
$^3$Microsoft Research Asia \\
  \texttt{sjtumaxb@sjtu.edu.cn, zhaohai@cs.sjtu.edu.cn,}\\ \texttt{\{yegong, nanduan\}@microsoft.com}, \texttt{Herbert.he@gmail.com}
  }

\makeatletter
\def\thanks#1{\protected@xdef\@thanks{\@thanks
        \protect\footnotetext{#1}}}
\makeatother
  
\begin{document}
\maketitle
\begin{abstract}
Based on the remarkable achievements of pre-trained language models in abstractive summarization, the copying mechanism has proved helpful by improving
the factuality, stability, and overall performance.
This work proposes \textbf{PROM}, a new \textbf{P}h\textbf{R}ase-level c\textbf{O}pying \textbf{M}echanism that enhances attention on $n$-grams, which can be applied to zero-shot summarization with pre-training.
PROM adds an indicator layer to explicitly pick up tokens in $n$-gram that can be copied from the source, and calculates an auxiliary loss for the copying prediction. 
Empirical studies show that PROM makes significant improvements in fine-tuning on benchmarks.
In the zero-shot setting, PROM is utilized in the self-supervised pre-training on raw corpora and provides new general baselines on a wide range of summarization datasets.
Further analysis shows that PROM performs more reasonable copying and contributes to faithfulness.
Our code is publicly available at \url{https://github.com/xbmxb/PROM}.
\end{abstract}

\section{Introduction}
The summarization task requires a model to comprehend an input passage and generate a summary.
An ideal summary covers the principal information of the source passage, shows consistency and faithfulness, and is fluent as human language \cite{pegasus, kryscinski-etal-2020-evaluating}.
Existing summarization strategies can be categorized into two main branches, abstractive \cite{rush-etal-2015-neural, nallapati-etal-2016-cnndm, zhou-etal-2017-selective, pegasus} and extractive \cite{nallapati2017summarunner, wang2019exploring, saggion2013automatic}.
Extractive summarization highly relies on extracting salient sentences from the source. 
Abstractive methods generate output sequences directly from the vocabulary, thus more flexible and closer to humans, but harder to control. 
Inspired by Transformer-series models \cite{vaswani2017attention}, abstractive methods are unified as conditioned seq2seq problem. 
Language models that have been pre-trained on large-scale corpora \cite{bart, 2020t5, pegasus, PALM, qi2020prophetnet} dominate this area of research.


The copying method represents a compromise of extraction and generation, alleviating the problems of inconsistency.
The consistency or faithfulness of abstractive summarization remains to be improved. 
Intrinsic reasons lie in the inherent imperfection of models, such as exposure bias \cite{liu-etal-2022-brio}, insufficient comprehension of the document \cite{wu-etal-2021-bass, dou-etal-2021-gsum}, while extrinsic reasons may be because the excessive confidence of the language model, leading to unfaithful summaries \cite{chen2022towards}. 
The copying method computes a copying distribution on the source sequence, and then aggregates the copying distribution and the language model distribution.  
Thus, unfamiliar tokens can be directly copied or ignored (e.g. new entities or out-of-vocabulary words) \cite{pointer, coconet}. 

Summarization also has to face the data bottleneck. On the one hand, high-quality summaries are usually human-generated \cite{nallapati-etal-2016-cnndm, xsum, wikihow}, but human writing shows diversity. On the other hand, language models require a large amount of data for supervised fine-tuning. 
Copying methods allow an alternative to picking up tokens from the source sequence, coping with expressions which the model is unfamiliar with. 
Intuitively, such an alternative agrees with the zero-shot situation or domain transfer.  

In this paper, we first propose a novel copying model with \textbf{PROM} (\textbf{P}h\textbf{R}ase-level c\textbf{O}pying \textbf{M}ethod) to enhance the attention of $n$-grams. Then we further propose a pre-training for zero-shot summarization. Transformer \cite{vaswani2017attention} is our backbone model, on which all our methods are implemented. 
Existing studies have indicated that (i) The language model contributes more to function words than entity words \cite{chen2022towards}; (ii) BART\cite{bart} has a tendency to copy sentences from the source \cite{zhang2022improving}. 
Thus, $n$-gram granularity can deserve much more attention for copying than random tokens. 
Instead of considering entities only \cite{chen2022towards, entity}, the enhancement needs to be extended to $n$-grams as they not only contain expressions of language but are also adaptive or common cross domains. 
Different from previous variants of copying module \cite{SAGCopy, entity, coconet}, we enhance the overlapped phrases and add an explicit loss. 
Experiments of supervised fine-tuning have proved the advantages of PROM compared to other copying methods.
Then we pre-train our model with PROM on raw corpora, to leverage PROM on zero-shot setting. 
Please note that the process of data construction leverages no characters of downstream testing tasks \cite{ted, wikitransfer, zhao-etal-2022-domain, brazinskas-etal-2020-unsupervised}.
Thus, our approach is supposed to be general across datasets or domains. 
Our pre-trained model contributes new zero-shot baselines on various widely used benchmarks, and achieves comparable scores against previous methods that are more domain-oriented. 

Contributions of our work are three folds:

(i) A copying method PROM that enhances copying attention of $n$-grams. Significant advantages have been proved with empirical studies on supervised fine-tuning.

(ii) A general self-supervised pre-training method that integrates PROM. Our pre-trained model provides new, widely ranged baseline scores in the zero-shot setting.

(iii) A detailed discussion from aspects of faithfulness, human evaluation, data bias, and comparison with large-scale language models.
Our model shows higher similarity to reference summaries, better factuality towards input passages, and scalability for various domains.

\section{Related Work}

\subsection{Copying Mechanism}
In seq2seq tasks, the copying mechanism allows the model to directly look at the source sequence during the generation. 
In addition to selecting a token from the vocabulary as the next one, picking a token from the source to copy is provided as an alternative.    
Building a bridge between the predicted token and the source sequence, the copying mechanism plays an important part in summarization.
COPYNET \cite{copynet} first introduced the copying mechanism into the seq2seq framework.
Simulating the rote memorization as humans will do, COPYNET shows effectiveness on short text summarization \cite{LCSTS} and single-turn dialogue response. 
Pointer-generator \cite{pointer} controls the copying mechanism by a calculated generation probability, leading to prominent progress on summarization benchmarks \cite{DBLP:conf/conll/NallapatiZSGX16, PALM}.

Recent studies bring up improved variants of copying.
Bottom-up attention \cite{bottomup} trains an extra content selector as a hard mask of copying distribution, where a threshold is used to filter source tokens. 
SAGCopy \cite{SAGCopy} deploys an attention graph for modeling relations among tokens, thus also improving the copying distribution. 
SeqCopyNet \cite{DBLP:conf/aaai/ZhouYWZ18} predicts an end position for the copying at each time to suggest a span to be extracted.
For multi-document summarization, a copying module helps with preserving details and handling rare tokens for respective docs \cite{DBLP:conf/acl/BrazinskasLT20}.
For better factual consistency \cite{DBLP:conf/emnlp/KryscinskiMXS20}, named entities in source sequence can be incorporated into the vocabulary, thus can be directly copied as an atomic unit \cite{entity}.
Coconet \cite{coconet} models the correlations of source tokens from semantic and positional perspectives and uses them to weight the copying distribution. The copying distribution is also fused with former values to make the generator aware of previous states.

\subsection{Low-Resource Summarization}
As high-quality summarization datasets are extremely expensive to acquire from the natural world \cite{DBLP:conf/emnlp/HuaW17,DBLP:journals/corr/abs-1908-11664}, the ability to adapt to multiple domains is expected on a well-trained model, specifically, in the few-shot or zero-shot setting. 
Pre-training is the most widely explored method for general improvements in multiple domains \cite{gururangan-etal-2020-dont}. 
Transformer-based language models that are pre-trained on large-scaled corpora \cite{bart, DBLP:journals/jmlr/RaffelSRLNMZLL20} achieve fluent and reasonable summaries for a wide range of datasets. Towards summarization task, specialized training objectives and training strategies even go a step further \cite{DBLP:conf/icml/ZhangZSL20, DBLP:conf/emnlp/QiYGLDCZ020, DBLP:journals/corr/abs-2208-09770}.

Recent studies carefully utilize task-oriented adaptation methods for better scores.
Adaptsum \cite{yu-etal-2021-adaptsum} fills the gap between pre-trained BART \cite{bart} and data of specific domains. The second pre-training methods provide benchmarks on low-resource domains like dialogue, social media, etc \cite{DBLP:conf/iclr/DinanRSFAW19, samsum, DBLP:conf/acl/RashkinSLB19, DBLP:conf/acl/KielaWZDUS18, DBLP:conf/naacl/KimKK19}.
TED \cite{ted} is pre-trained on pseudo summaries constructed from news and transferred within news domains only, using the methods of theme modeling and denoising. 
Subsequently, WikiTransfer \cite{DBLP:conf/naacl/FabbriHLLGJRM21} simulates features of the target tasks from aspects of extractive diversity and compression ratio. This task-oriented method leads to significant improvement, but also requires specialized pseudo data and models for each target task.
Prompt learning \cite{DBLP:conf/coling/LiuGBLHHC22} is also introduced into the few-shot summarization. Following prompt pre-training, the fine-tuning uses only 300 samples and balances the effectiveness-efficiency trade-off.
This work combines the proposed novel copying method PROM with zero-shot summarization by a general pre-training, achieving higher scores than mainstream baselines.
\begin{figure*}[t]
		\centering
		\includegraphics[width=\textwidth]{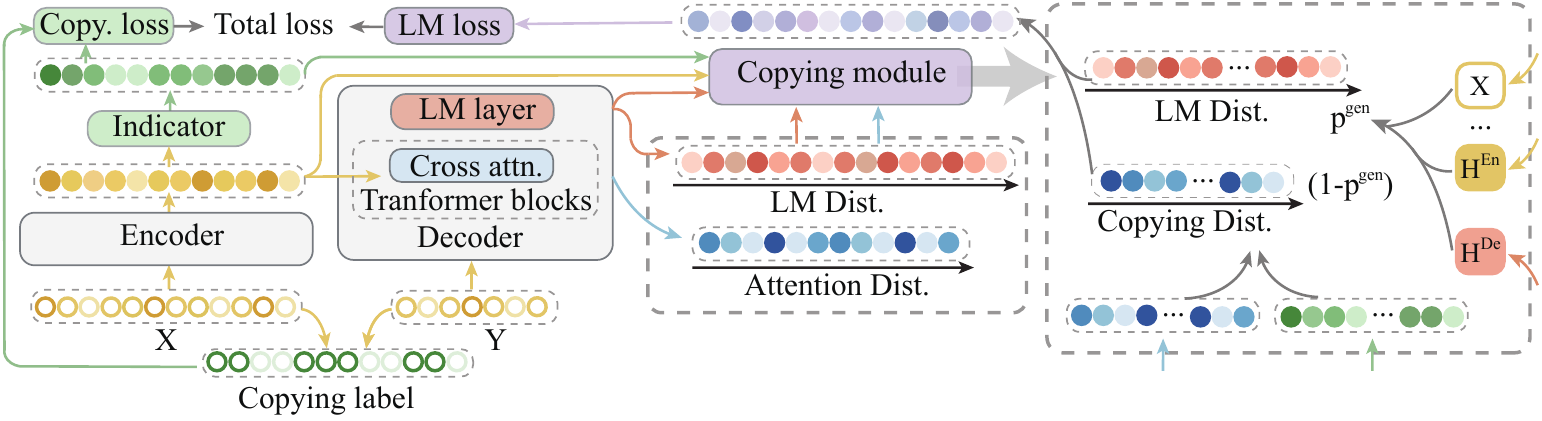}
		\caption{\label{model} Overview of the proposed PROM. The left part shows the architecture of our model consisting of the Encoder, Decoder, and Copying module, while the right part shows a closer look at the Copying module.}
\end{figure*}
More recently, large language models (LLMs) have shown impressive zero- and few-shot abilities \cite{ouyang2022training, brown2020language, chowdhery2022palm} and scalability. Input an instruction, the LLM can recognize the intent and give human-like responses. Whereas, the larger scale consumes heavier computational resources. Comparisons are presented in our analyses.

\section{Methodology}
In this section, after defining the summarization task, we introduce PROM and our pre-training with PROM for zero-shot setting. 
Summarization can be formulated as a seq2seq task, where an output summary $Y = [y_0, y_1, y_2, ..., y_t]$ is expected given a source sequence (article, news, log) $X = [x_0, x_1, x_2, ... , x_s]$.
In the following, the reference summary is denoted as $Y$ while the predicted summary is denoted as $Y_{prd}$.
An overview of our model is shown in Figure \ref{model}. PROM follows the Transformer-based encoder-decoder framework and utilizes the copying module to encourage precise and reasonable copying.

\subsection{PROM}
\subsubsection{Backbone}
Transformer-based seq2seq models stack attention layers for encoding and decoding, and produce subsequent tokens from vocabulary auto-regressively.
This process can be denoted as
\begin{equation}
\setlength{\abovedisplayskip}{7pt}
\setlength{\belowdisplayskip}{7pt}
\begin{split}
& {H}_{t}^{En} = {F}_{En} {( } {X} { )}, \\
& {H}_{t}^{De} = {F}_{De} {( } {H}_{t}^{En} ,  { [} {y}_{0} { : } {y}_{t-1} {] )},\\
& {P}_{t}^{vocab} = {lm(} {H}_{t}^{De} {)},\\
& \mathcal{L}_{summ} = \sum_{t} {CE} {( } {P}_{t}^{vocab} , {y}_{t} { )} ,
\end{split}
\label{transformer}
\end{equation}
where $lm$ is the feed-forward language model layer. 
The loss function is the cross entropy $CE$ between the distribution on vocabulary ${P}^{vocab}$ and label $Y$.

\subsubsection{Copying with Phrase Enhancement}
\label{312}
With the assistance of the copying mechanism, the predicted distribution becomes a combination of logits on both vocabulary and source tokens \cite{copynet, pointer, coconet}. 
Intuitively, they respectively stand for generating from vocabulary and copying from the source.
\begin{equation}
\setlength{\abovedisplayskip}{7pt}
\setlength{\belowdisplayskip}{7pt}
\begin{split}
& {P}_{t} {(} {w} {)} = {p}_{t}^{gen} {P}_{t}^{vocab} {(} {w} {)}  + {p}_{t}^{copy} {P}_{t}^{copy} {(} {w} {)}, \\
& {p}^{copy} = {1 - } {p}^{gen}, \\
\end{split}
\label{pointer}
\end{equation}
where $w$ is for some certain sub-word (token) and $FC$ is short for fully connected layer. $\textit{p}_{t}^{gen}$ and $\textit{P}_{t}^{copy}$ depend on interactions with the source sequence, mostly by cross attention mechanism in existing work.

However, it is shown that PrLMs may have the over-copying problem with summarization.
In consideration of this issue, the copying probability of each source token can be explicitly modeled to make the copying more reasonable.
First, we make pseudo-labels for copying in the phrase granularity.
Tokens in overlapped $n$-grams between the source articles and the reference summaries are tagged to be copied.
An $n$-token length window goes through the source sequence $X$, extracting each $n$-gram sequence from $X$. If the same $n$-gram exists in the target sequence $Y$, tokens in that window are labeled. $C$ denotes the label sequence and can be formulated as
\begin{equation}
\setlength{\abovedisplayskip}{7pt}
\setlength{\belowdisplayskip}{7pt}
\begin{split}
& {C} = {[} {c}_{0} {, } {c}_{1} {, } {c}_{2} {, } \dots {, } {c}_{i} {, } \dots {, } {c}_{s}  {]} , \\ 
& {c}_{i} = 
\begin{cases}
1 {,}& {[} {c}_{i-j} {, } { c}_{i-j+n-1} {] \quad in \quad} { Y} \\
0 {,}& {otherwise}\\
\end{cases} , \\
&  { n \geq } {0 , 0 } {\leq j \le n}.
\end{split}
\label{copylabel}
\end{equation}
We present an example of bi-gram tagging in Table \ref{tab:eg4indicator}.
Labeling $n$-grams makes a larger scope than named entities (e.g. \textit{at the Ecuadorian Embassy}), and leaves other tokens like function words to the language model (e.g. the first \textit{is}, the third word in the summary).
(Appendix \ref{appa} shows details of this example with analysis.)
This method makes intuitive sense as 
(i) Sequential overlaps are selected and the odd tokens are filtered;
(ii) Overlapped $n$-grams are often more meaningful than uni-grams. Thus, phrases are enhanced in our copying module, like named entities or grammar patterns.
\begin{table}[bth]
	\centering\small
        \resizebox{\linewidth}{!}{
	\begin{tabularx}{0.9\linewidth}{X}
		\toprule
		\textbf{Article:}   \\
		 Hollywood actor John Cusack is the latest supporter to visit WikiLeaks founder Julian Assange in his continued stay at the Ecuadorian Embassy... Assange has avoided being extradited to Sweden by taking shelter in the Ecuadorean Embassy in London since 2012...
		\\
		\midrule
		\textbf{Summary:} \\
            \textcolor{blue}{Hollywood actor} is \textcolor{blue}{latest supporter to visit WikiLeaks founder} Assange. Pictured arriving \textcolor{blue}{at the Ecuadorian Embassy} where Assange is staying. Assange is avoiding extradition \textcolor{blue}{to Sweden by taking shelter in} embassy. \\
		\bottomrule
	\end{tabularx}}
	\caption{An example for the copying indicator. The labeled bi-grams (sub-word token) are in blue.}
        \label{tab:eg4indicator}
\end{table}

Then we add an indicator layer on the top of the encoder.
It is a linear module that takes the encoder hidden states as input and predicts the probability of copying $H_C$. 
Then a cross-entropy loss can be computed for the probability $H_C$ and the copying label $C$. 
\begin{equation}
\setlength{\abovedisplayskip}{7pt}
\setlength{\belowdisplayskip}{7pt}
\begin{split}
& {H}_{C} = sigmoid {(} FC{( } {H}_{t}^{En}  {))} , \\ 
& \mathcal{L}_{copy} = CE {(} { H}_{C} {, } {C } {)}.
\end{split}
\label{indicator}
\end{equation}
In our copying module, the copying prediction $H_C$ is integrated to facilitate the copying distribution.
Source tokens that have high copying probabilities should be more likely to be selected.
\begin{equation}
\setlength{\abovedisplayskip}{7pt}
\setlength{\belowdisplayskip}{7pt}
\begin{split}
& {a}_{C} = sigmoid {(} FC {( } {H}_{C} { , } {a}  { ))} , \\ 
& {\tilde{P}}_{t}^{copy} {(} {w} {)} = \sum_{w} {a}_{C}, \\
& {p}^{gen} = sigmoid {(} FC {( } {a} \cdot {H}^{En} { , } {H}^{De} { , } {X} { ))} , \\  
& {p}^{copy} = {1 - } {p}^{gen}, \\
& {\tilde{P}}_{t} {(} {w} {)} = {p}_{t}^{gen} {P}_{t}^{vocab} {(} {w} {)} { + } {p}_{t}^{copy} {\tilde{P}}_{t}^{copy} {(} {w} {)},
\end{split}
\label{copygate}
\end{equation}
where $a$ is cross attention scores of input tokens, and $a_C$ is the scores integrated with $H_C$.
In terms of the training objective, we use the mainstream cross-entropy loss $\mathcal{L}_{summ}$ and also add $\mathcal{L}_{copy}$.
\begin{equation}
\setlength{\abovedisplayskip}{7pt}
\setlength{\belowdisplayskip}{7pt}
\begin{split}
& \mathcal{L}_{summ} = \sum_{t} CE {(} \textit{ $\tilde{P}$}_{t} {, } {y}_{t} { )}, \\
& \mathcal{L} = \mathcal{L}_{summ} { + } \lambda \mathcal{L}_{copy}.
\end{split}
\label{loss}
\end{equation}
Two training strategies are tried on our model:
(i) multi-task method: use the total loss $\mathcal{L}$ as the training objective.
(ii) two-stage method: firstly train the copying indicator for several steps, $\mathcal{L}_{copy}$ as the training objective, and then add up $\mathcal{L}_{summ}$ for more steps, $\mathcal{L}$ as the training objective.

\section{Pre-training for Few-shot Setting}\label{pt}
The copying method builds a bridge for input tokens over the deep stacks of transformer layers, thus helping with OOD \cite{copynet, coconet} and improving the stiffness \cite{chen-etal-2020-cdevalsumm}. 
Naturally, humans can deal with unfamiliar words by copying without truly understanding them if they infer that the words are important. 
Motivated by this, we apply our PROM to zero-shot setting.

We propose to pre-train with PROM on the self-supervised objective to leverage our copying module for performance improvement and generalization on the zero-shot setting. 
Our self-supervised training dataset is constructed from corpora.
Formally, let $D$ denote a natural passage in the corpora, which consists of several sentences $D = \{d_0, d_1, d_2, \dots d_{\hat{s}} \}$. 
$D$ is processed into pseudo document-summary pairs $(\hat{X}, \hat{Y})$ in the following two ways.

(\romannumeral1) $D_{nat}$: 
Given a passage $D$, we calculate important score $ Score ( i )$ for each sentence $d_i$.
The $m\%$ top-scoring sentences are selected and deleted from $D$.
Then a pseudo document-summary pair $(\hat{X}, \hat{Y})$ is generated, where $\hat{X}$ is the selected sentences and $\hat{Y}$ is the remaining passage.
Both $\hat{X}$ and $\hat{Y}$ keep the original order. 
This method follows gap sentence generation (GSG) \cite{pegasus} but uses Extractive Fragments Density (EFD) \cite{grusky-etal-2018-newsroom} as the importance score,
\begin{equation}
\setlength{\abovedisplayskip}{7pt}
\setlength{\belowdisplayskip}{7pt}
\begin{split}
& \operatorname{EFD}(x, y)=\sum_{f \in \mathcal{F}(x, y)}|f|^2 / |x|, \\
& Score(i) = \operatorname{EFD} ( d_i, D \setminus \{d_i\}),
\end{split}
\label{pseudodata}
\end{equation}
where $\mathcal{F}(x, y)$ is overlapped fragments of sequences $x, y$.

(\romannumeral2) $D_{chunk}$: 
However, natural articles vary widely in length, and long articles may be truncated and wasted to fit the model width.
Thus, we augment our data by setting maximum and minimum limitations on the document sentence number.
Passages are chunked by the maximum, and those shorter than the minimum are discarded. 
Then the same selection is performed on the chunks as (\romannumeral1).

To be consistent with our copying method, we loosely control the extractiveness level of the pre-training data. The data is filtered by a minimum EFD $min_{EFD}$. 
Finally, the pre-training dataset is the filtered union of $D_{nat}$ and $D_{chunk}$.
The training objective is the total loss $\mathcal{L}$ in equation \ref{loss}.

Our model is trained on the pre-training dataset and evaluated on a wide range of downstream tasks in a zero-shot way. 
Please note that downstream tasks have different characters in the document and summary length, compression ratio, the proportion of novel tokens, etc.
But no downstream task-specialized processing is performed on the pre-training data \cite{ted, wikitransfer}.
Thus no information for target data is available, 
which is against our goal of the generalized zero-shot method.

Lead bias is a common feature of summarization datasets that is often utilized \cite{wikitransfer, yang2020ted}.
It is reasonable that the front part contains more primary information of a document. 
For comparison and analysis, we also create a lead-biased pseudo dataset by extracting the first $\hat{m}$ sentences as summaries and leaving the rest as documents. 

\begin{table}[tb]
	\centering\small
	{
	\begin{tabular}{p{1.1cm}p{1.1cm}p{1.1cm}p{1.4cm}p{0.5cm}}
	\toprule
	
	  \multirow{2}{*}{\textbf{Data}}  & \multirow{2}{*}{\textbf{Genre}} & \makebox[0.9cm][c]{\textbf{Size}} & \makebox[1.6cm][c]{\textbf{Len.}(words)}  & \multirow{2}{*}{\textbf{Ratio}} \\
        & & \makebox[0.9cm][c]{train/test} & \makebox[1.6cm][c]{Doc./Sum.}  & \\
        \midrule
    \multicolumn{5}{c}{\textit{Summerization Datasets}}\\
    CNN/DM  & news &287k/11k & 682/54  & 14.03 \\
    NYT & news  & 146k/17k &990/79 & 13.21   \\
    BillSum & bill  & 19k/2k &1219/174 & 6.25   \\
    WikiHow & instruction  & 168k/6k &580/62 & 10.96  \\
    arXiv & science & 203k/6k &4938/220 & 35.58  \\
    XSum & news & 204k/11k &361/21 & 18.25  \\
	\bottomrule
	\end{tabular}
	}
	\caption{Statistics and characters of datasets.}
	\label{data}
\end{table}

\section{Experiments}

\subsection{Dataset}\label{datasets}

\subsubsection{Summarization data}
\begin{figure}[htbp]     
    \centering      
        \begin{minipage}{0.49\textwidth}    
            \centering      
            \includegraphics[width=\textwidth]{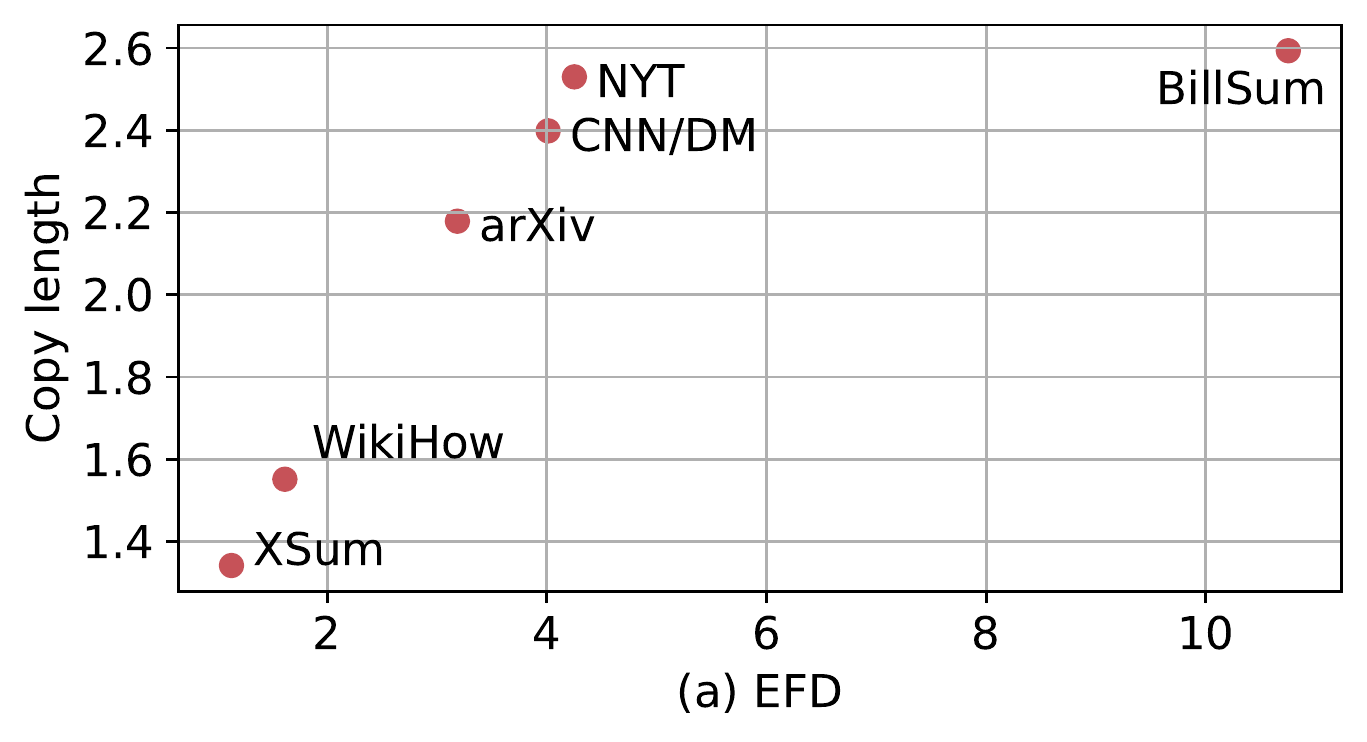}
        \end{minipage}            
        \begin{minipage}{0.49\textwidth}
            \centering      
            \includegraphics[width=\textwidth]{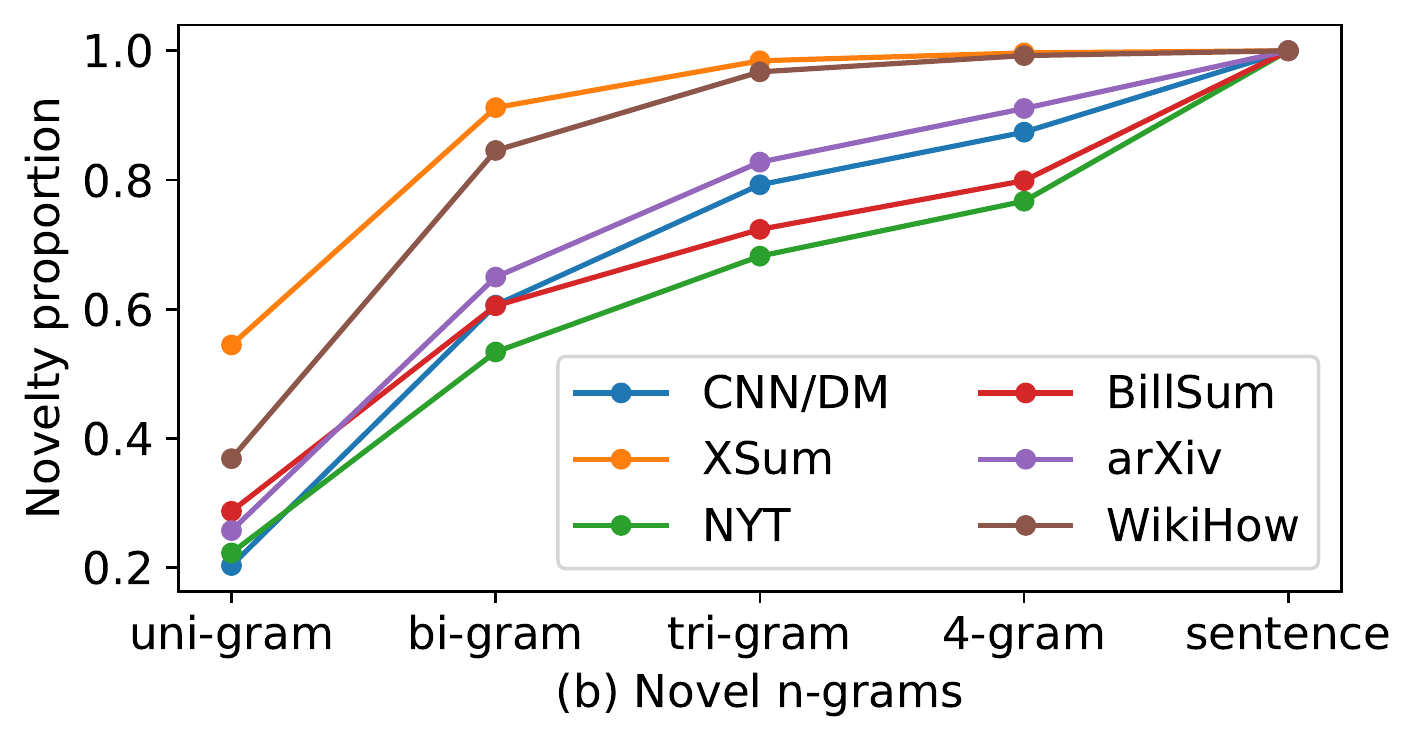}      
        \end{minipage}     
    \caption{\label{ext} (a) Extractive Fragments Density \& Copy Length. (b) $n$-gram Novelty.}
\end{figure}

For summarization tasks, our empirical studies rely on a series of datasets that are diverse in genre, size, length, and also extractiveness or abstractiveness level. 
The statistics and characters are listed in Table \ref{data} and Figure \ref{ext}.

\textbf{CNN/DailyMail} \cite{nallapati-etal-2016-cnndm} 
consists of 93k articles from CNN News and 220k articles from Daily Mail News. We use the non-anonymized version \cite{pointer} by default.

\textbf{New York Times} is derived from published news from New York Times, whose annotation is conducted by experts or hand-verified. We obtain over 174k examples from the corpus for experiments.

\textbf{BillSum} \cite{kornilova-eidelman-2019-billsum} contains 22,218 US Congressional bills. The reference summaries are human-written from the Congressional Research Service.

\textbf{WikiHow} \cite{wikihow} releases 230k summarization examples constructed from WikiHow.com, an online knowledge base. The articles are human-written instructions and summaries are combined subtitles.

\textbf{arXiv} \cite{arxiv} is a collection of 113k scientific papers on arXiv.org. Summaries are annotated as the abstracts of the papers while the remaining are the documents to be summarized. It is a representative long-document summarization dataset.
 
\textbf{XSum} \cite{xsum} is derived from BBC articles annotated with human-written single-sentence summaries. It consists of 227k samples that are diverse in topics.

\noindent\textbf{Extractiveness \& Abstractiveness Level.}
For the copying method, the level of extractiveness \& abstractiveness is an important feature of datasets.
To characterize this, three metrics are illustrated in Figure \ref{ext}.
\textbf{(i) Extractive Fragments Density} \cite{grusky-etal-2018-newsroom} (Eq. \ref{pseudodata}) describes the density of shared fragments, thus reflects how well the summary can be regarded as a series of extractions.
\textbf{(ii) Copy Length} \cite{grusky-etal-2018-newsroom, chen-etal-2020-cdevalsumm} computes the average length of shared fragments, indicating the tendency of continuous copying.
\textbf{(iii) $n$-gram Novelty} \cite{pointer, sharma-etal-2019-bigpatent} computes the proportion of $n$-grams that exists in the summary but not in the article. 
It can be observed that XSum and WikiHow are significantly more abstractive than others. 
BillSum encourages extraction the most, with the largest EFD and Copy Length. 
NYT is the most conservative for novel phrases.
Overall, the tendency ranking from abstractive generation to extraction is \textit{XSum $>$ WikiHow $>$ arXiv $>$ CNN/DM $>$ NYT $>$ BillSum}.

\subsubsection{Pre-training Corpora}\label{subset}
For pre-training, we use the large-scale corpora of ProphetNet \cite{qi2020prophetnet}.
It consists of 160Gb English articles collected from news, stories, and web text, and there is no overlap between the corpora and the above summarization datasets.
In our implementation, we first conduct experiments on a 29G subset due to the limited resources. 
Articles of different genres (news, stories, web text) keep almost the same proportions in the subset as the full-size corpora.

\subsection{Experimental Setup}
Our empirical study consists of two steps. 
We first prove the effectiveness of PROM by fine-tuning on summarization datasets. 
Then we apply PROM to pre-training and evaluate the zero-shot performance, which verifies our motivation to enhance zero-shot summarization with copying.

\subsubsection{Fine-tuning Setting}
Our model is first fine-tuned to evaluate the effectiveness of PROM.
Among the six benchmarks, we choose the most commonly used CNN/DM to compare with previous work (Table \ref{modelchange}).
We also choose datasets with different characters WikiHow (more abstractive), and arXiv (long documents) to prove the scalability (Table \ref{wikiar}).

The previous studies to compare contain commonly used methods and summarization systems related to copying, for example, Point-Generator \cite{pointer}, SAGCopy \cite{SAGCopy}, and Bottom-Up \cite{bottomup}. Notably, CoCoNet \cite{coconet} is the previous SOTA of copying methods. These methods all provide results on CNN/DM.
For WikiHow and arXiv, the baselines are Lead \cite{pointer}, Point-Generator \cite{pointer}, PEGASUS \cite{pegasus}, and plain BART.

\begin{table}[tbh]
	\centering\small
	{\begin{tabular}{p{4.4cm}p{0.4cm}p{0.4cm}p{0.6cm}}
		\toprule
		\textbf{Model} & \textbf{R$_{1}$} &\textbf{R$_{2}$} &\textbf{R$_{L}$}\\ 
		\midrule
		\multicolumn{3}{c}{\emph{Previous work}} \\
            Lead \cite{pointer}& 40.34 & 17.70 & 36.57 \\
            Point-Generator \cite{pointer}& 39.53 & 17.28 & 36.38 \\
            DRM \cite{DBLP:conf/iclr/PaulusXS18}& 39.87 & 15.82 & 36.90 \\
            Bottom-Up \cite{bottomup}& 41.22 & 18.68 & 38.34 \\
            S2S-ELMo \cite{edunov-etal-2019-pre} & 41.56 & 18.94 & 38.47 \\
            DCA \cite{celikyilmaz-etal-2018-deep}& 41.69 & 19.47 & 37.92 \\
            BERTSUMEXTABS & 42.13 & 19.60 & 39.18 \\
            \cite{liu-lapata-2019-text} &&&\\
            MASS \cite{song2019mass} & 42.12 & 19.50 & 39.01 \\
            SAGCopy \cite{SAGCopy} & 42.53 & 19.92 & 39.44 \\
            UniLM \cite{DBLP:conf/nips/00040WWLWGZH19}& 43.33 & 20.21 & 40.51 \\
            ProphetNet-16G \cite{qi2020prophetnet}& 43.68 & 20.64 & 40.72 \\
            BART$_{large}$ (Reported) & 44.16 & 21.28 & 40.90 \\
            T5 \cite{2020t5} & 43.52 & 21.55 & 40.69 \\
            PEGASUS \cite{pegasus} & 44.17 & 21.47 & 41.11 \\
            ProphetNet \cite{qi2020prophetnet}& 44.20 & 21.17 & 41.30 \\
            PALM \cite{PALM}& 44.30 & 21.12 & 41.41 \\
            BART+SAGCopy \cite{coconet}&44.31 &21.35 &41.00 \\
            CoCoNet \cite{coconet}&44.39 &21.41 &41.05 \\
            CoCoPretrain \cite{coconet}&44.50 &21.55 &41.24 \\
            \midrule
            \multicolumn{3}{c}{\emph{Our implementations}} \\
            BART$_{large}$ (Our) & 43.79 & 21.20 & 40.70 \\
            BART$_{large}$ (Fb) & 44.11 & 21.08 & 40.91 \\
            BART$_{large}$+Pointer-Gen. & 44.11 & 21.27 & 40.98 \\
            PROM & 44.47 & 21.59 & 41.32 \\
            PROM$_{two\_stage}$ & 44.35 & 21.61 & 41.19 \\
            PROM$_{pre-train}$ & \textbf{44.59} & \textbf{21.66} & \textbf{41.46} \\
		\bottomrule
	\end{tabular}
	}
        \caption{ROUGE $\operatorname{F}_1$ scores on CNN/DM.}
	\label{modelchange}
\end{table}

\begin{table}[htb]
	\centering\small
	{\begin{tabular}{p{1.7cm}p{0.4cm}p{0.4cm}p{0.5cm}p{0.4cm}p{0.4cm}p{0.4cm}}
		\toprule
		\multirow{2}{*}{\textbf{Model}} & \multicolumn{3}{c}{\textbf{WikiHow}} &\multicolumn{3}{c}{\textbf{arXiv}} \\ 
            & \makebox[0.4cm][c]{\textbf{R$_{1}$}} & \textbf{R$_{2}$} & \textbf{R$_{L}$} & \textbf{R$_{1}$} & \textbf{R$_{2}$} & \textbf{R$_{L}$}\\
		\midrule
            Lead & \makebox[0.4cm][c]{24.97} &\makebox[0.4cm][c]{5.83} &\makebox[0.4cm][c]{23.24}  & \makebox[0.4cm][c]{28.05} & \makebox[0.4cm][c]{6.63} & \makebox[0.4cm][c]{17.72}\\
            Pointer-Gen.  & \makebox[0.4cm][c]{28.53} &\makebox[0.4cm][c]{9.23} &\makebox[0.4cm][c]{26.54}  & \makebox[0.4cm][c]{32.06} & \makebox[0.4cm][c]{9.04} & \makebox[0.4cm][c]{25.16}  \\
           
            PEGASUS  & \makebox[0.4cm][c]{43.06} &\makebox[0.4cm][c]{19.71} &\makebox[0.4cm][c]{--}   & \makebox[0.4cm][c]{44.70} & \makebox[0.4cm][c]{17.27} & \makebox[0.4cm][c]{--} \\ 
             BART$_{large}$  & \makebox[0.4cm][c]{45.22} &\makebox[0.4cm][c]{20.13} &\makebox[0.4cm][c]{43.73}  & \makebox[0.4cm][c]{45.18} & \makebox[0.4cm][c]{16.87} & \makebox[0.4cm][c]{39.42}  \\
            PROM  & \makebox[0.4cm][c]{45.57} &\makebox[0.4cm][c]{20.53} &\makebox[0.4cm][c]{44.09} & \makebox[0.4cm][c]{45.24} & \makebox[0.4cm][c]{16.95} & \makebox[0.4cm][c]{39.38}  \\
            \bottomrule
	\end{tabular}
	}
        \caption{ROUGE $\operatorname{F}_1$ scores on datasets WikiHow and arXiv to show the scalability of PROM.}
	\label{wikiar}
\end{table}

We implement the following settings. In Table \ref{modelchange}, three results of BART$_{large}$ are presented. ``\textit{Reported}'' denotes scores reported in \citet{bart}. ``\textit{Fb}'' denotes our test results of the released \textit{facebook/bart-large-cnn}. ``\textit{Our}'' denotes test results of our fine-tuning on \textit{facebook/bart-large}. ``\textit{BART$_{large}$+Pointer-Gen.}'' denotes integration of BART and copying, i.e., $\lambda$ in Eq. 6 is 0. For PROM implementations, ``\textit{two-stage}'' is the second training strategy in Section \ref{312}. ``\textit{pre-train}'' means fine-tuning after our pre-training.

Our implementation is based on \textit{Transformers} \cite{wolf-etal-2020-transformers}.
The model is initialized with the official weights of pre-trained BART$_{large}$. 
The hidden size is 1024 by default. 
Learning rate is set to \{1e-5, 3e-5, 5e-5\}, while batch size is set to \{36, 48, 64\}. 
The beam search during decoding is in size of 4. For simplicity, $n$ is set to 2, and $\lambda$ is set to 1.
The epoch number is set to \{4, 6, 8\}, and the best checkpoint is selected by the validation set. More details are shown in the Appendix. 
Appendix \ref{appb} shows more experimental details.

\begin{table}[tbh]
	\centering\small
	{\begin{tabular}{p{1.7cm}p{0.4cm}p{0.4cm}p{0.5cm}p{0.4cm}p{0.4cm}p{0.4cm}}
		\toprule
		{\textbf{Model}} & \makebox[0.4cm][c]{\textbf{R$_{1}$}} & \textbf{R$_{2}$} & \textbf{R$_{L}$} & \textbf{R$_{1}$} & \textbf{R$_{2}$} & \textbf{R$_{L}$} \\ 
            \midrule
            & \multicolumn{3}{c}{\textbf{CNN/DM}} &\multicolumn{3}{c}{\textbf{XSum}}\\
		\midrule
            Lead & \makebox[0.4cm][c]{40.34} &\makebox[0.4cm][c]{17.70} &\makebox[0.4cm][c]{36.57}  &\makebox[0.4cm][c]{16.30} &\makebox[0.4cm][c]{1.60} &\makebox[0.4cm][c]{12.00}\\
            TED  & \makebox[0.4cm][c]{38.38} &\makebox[0.4cm][c]{16.49} &\makebox[0.4cm][c]{35.08}  & \makebox[0.4cm][c]{--} & \makebox[0.4cm][c]{--} & \makebox[0.4cm][c]{--}  \\
            WikiTransfer & \makebox[0.4cm][c]{40.14} &\makebox[0.4cm][c]{17.71} &\makebox[0.4cm][c]{36.66}   &\makebox[0.4cm][c]{31.85} & \makebox[0.4cm][c]{10.44} & \makebox[0.4cm][c]{23.75} \\
            \quad \text{w/o bin}  & \makebox[0.4cm][c]{39.11} &\makebox[0.4cm][c]{16.98} &\makebox[0.4cm][c]{35.66}   &\makebox[0.4cm][c]{22.78} & \makebox[0.4cm][c]{5.66} & \makebox[0.4cm][c]{17.16} \\ 
            \quad \text{w/ GSG} & \makebox[0.4cm][c]{37.62} &\makebox[0.4cm][c]{15.15} &\makebox[0.4cm][c]{34.21}  &\makebox[0.4cm][c]{29.95} & \makebox[0.4cm][c]{9.37} & \makebox[0.4cm][c]{21.78} \\ 
            PEGASUS  & \makebox[0.4cm][c]{32.90} &\makebox[0.4cm][c]{13.28} &\makebox[0.4cm][c]{29.38}   &\makebox[0.4cm][c]{19.27} & \makebox[0.4cm][c]{3.00} & \makebox[0.4cm][c]{12.72} \\ 
            
            PROM$_\textup{subset}$  & \makebox[0.4cm][c]{37.34} &\makebox[0.4cm][c]{15.26} &\makebox[0.4cm][c]{33.53}  &\makebox[0.4cm][c]{22.92} & \makebox[0.4cm][c]{6.30} & \makebox[0.4cm][c]{17.53} \\
            PROM$_\textup{full}$   & \makebox[0.4cm][c]{37.87} &\makebox[0.4cm][c]{15.91} &\makebox[0.4cm][c]{34.16} & \makebox[0.4cm][c]{22.96} & \makebox[0.4cm][c]{6.05} & \makebox[0.4cm][c]{17.78} \\
            
		\midrule
            \multirow{2}{*}{} & \multicolumn{3}{c}{\textbf{NYT}} &\multicolumn{3}{c}{\textbf{WikiHow}} \\ 
            \midrule
            Lead &  \makebox[0.4cm][c]{35.50} & \makebox[0.4cm][c]{17.20} & \makebox[0.4cm][c]{32.00}  &\makebox[0.4cm][c]{24.97} &\makebox[0.4cm][c]{5.83} &\makebox[0.4cm][c]{23.24} \\
            TED   & \makebox[0.4cm][c]{35.03} & \makebox[0.4cm][c]{16.57} & \makebox[0.4cm][c]{31.96} & \makebox[0.4cm][c]{--} & \makebox[0.4cm][c]{--} & \makebox[0.4cm][c]{--}  \\
            PEGASUS & \makebox[0.4cm][c]{--} & \makebox[0.4cm][c]{--} & \makebox[0.4cm][c]{--} & \makebox[0.4cm][c]{22.59} & \makebox[0.4cm][c]{6.10} & \makebox[0.4cm][c]{14.44} \\ 
            
            PROM$_\textup{subset}$  & \makebox[0.4cm][c]{36.37} & \makebox[0.4cm][c]{16.45} & \makebox[0.4cm][c]{28.88} &  \makebox[0.4cm][c]{25.40} & \makebox[0.4cm][c]{6.04} & \makebox[0.4cm][c]{23.55} \\
            PROM$_\textup{full}$  & \makebox[0.4cm][c]{36.95} &\makebox[0.4cm][c]{17.21} &\makebox[0.4cm][c]{29.42}  & \makebox[0.4cm][c]{25.90} & \makebox[0.4cm][c]{6.37} & \makebox[0.4cm][c]{24.08}\\
            \midrule
            \multirow{2}{*}{} & \multicolumn{3}{c}{\textbf{BillSum}} &\multicolumn{3}{c}{\textbf{arXiv}} \\ 
            \midrule
            Lead & \makebox[0.4cm][c]{21.09} &\makebox[0.4cm][c]{7.66} &\makebox[0.4cm][c]{18.18}  &\makebox[0.4cm][c]{26.46} &\makebox[0.4cm][c]{6.28} &\makebox[0.4cm][c]{22.75}\\
            PEGASUS  &  \makebox[0.4cm][c]{41.02} & \makebox[0.4cm][c]{17.44} & \makebox[0.4cm][c]{25.24}  & \makebox[0.4cm][c]{28.05} & \makebox[0.4cm][c]{6.63} & \makebox[0.4cm][c]{17.72} \\ 
            
            PROM$_\textup{subset}$   & \makebox[0.4cm][c]{39.77} & \makebox[0.4cm][c]{13.88} & \makebox[0.4cm][c]{33.05} & \makebox[0.4cm][c]{34.22} & \makebox[0.4cm][c]{9.29} & \makebox[0.4cm][c]{29.72} \\
            PROM$_\textup{full}$  & \makebox[0.4cm][c]{40.05} &\makebox[0.4cm][c]{14.66} &\makebox[0.4cm][c]{33.34}  & \makebox[0.4cm][c]{34.90} & \makebox[0.4cm][c]{9.71} & \makebox[0.4cm][c]{30.43} \\
            \bottomrule
	\end{tabular}
	}
        \caption{Zero-shot ROUGE $\operatorname{F}_1$ results of pre-training with PROM. Our results on both subset (Section \ref{subset}) and full corpora are reported.}
	\label{pretrain}
\end{table}

\subsubsection{Zero-shot Setting}
In experiments of pre-training for zero-shot setting, our model with PROM is pre-trained on the self-supervised data based on the corpora. Then, the model is tested on all of the six datasets.

The baselines for zero-shot results are Lead \cite{pointer}, TED \cite{ted}, PEGASUS \cite{pegasus}, and WikiTransfer \cite{wikitransfer}. Three settings of WikiTransfer are presented, the best setting, without extractiveness filtering, (w/o bin), and using GSG for summary sentence choice (w/ GSG). Please note that WikiTransfer leverages data features and others are strict zero-shot settings.

We load BART$_{large}$ checkpoints for continuous pre-training on our constructed data. Learning rate is set to \{1e-6, 1e-5\}, while batch size is \{80, 128\}. The total epoch number is \{2, 4, 6\}. Other hyperparameters are the same as those in fine-tuning. Then the model is tested on downstream datasets without fine-tuning. Results are shown in Table \ref{pretrain}.

\section{Analysis}

\subsection{Main Results}\label{rougescore}
ROUGE $\operatorname{F}_1$ scores are reported as main results in Table \ref{modelchange}, Table \ref{wikiar}, and Table \ref{pretrain}.

In Table \ref{modelchange}, our model surpasses all previous copying methods.
PROM leads to significant improvements on each ROUGE score, especially on ROUGE-2. 
Two-stage training leads to a higher ROUGE-2 score but lower ROUGE-1 and ROUGE-L. 
Fine-tuned after our pre-training, PROM$_{pre-train}$ achieves even better performance compared to all related baselines.
Table \ref{wikiar} shows the scalability on WikiHow, a more abstractive task, and arXiv, a long-document task. Gains on CNN/DM and WikiHow are larger than on longer sequences.

In terms of the zero-shot results in Table \ref{pretrain}, PROM surpasses PEGASUS on most of the datasets except BillSum. This may be because BillSum agrees with over-copying.
Compared to methods that utilize domain features, PROM gets lower but comparable scores.
Thus, PROM provides a new summarization baseline under the strict zero-shot setting.

\begin{figure}[hbt]
		\centering
		\includegraphics[width=0.49\textwidth]{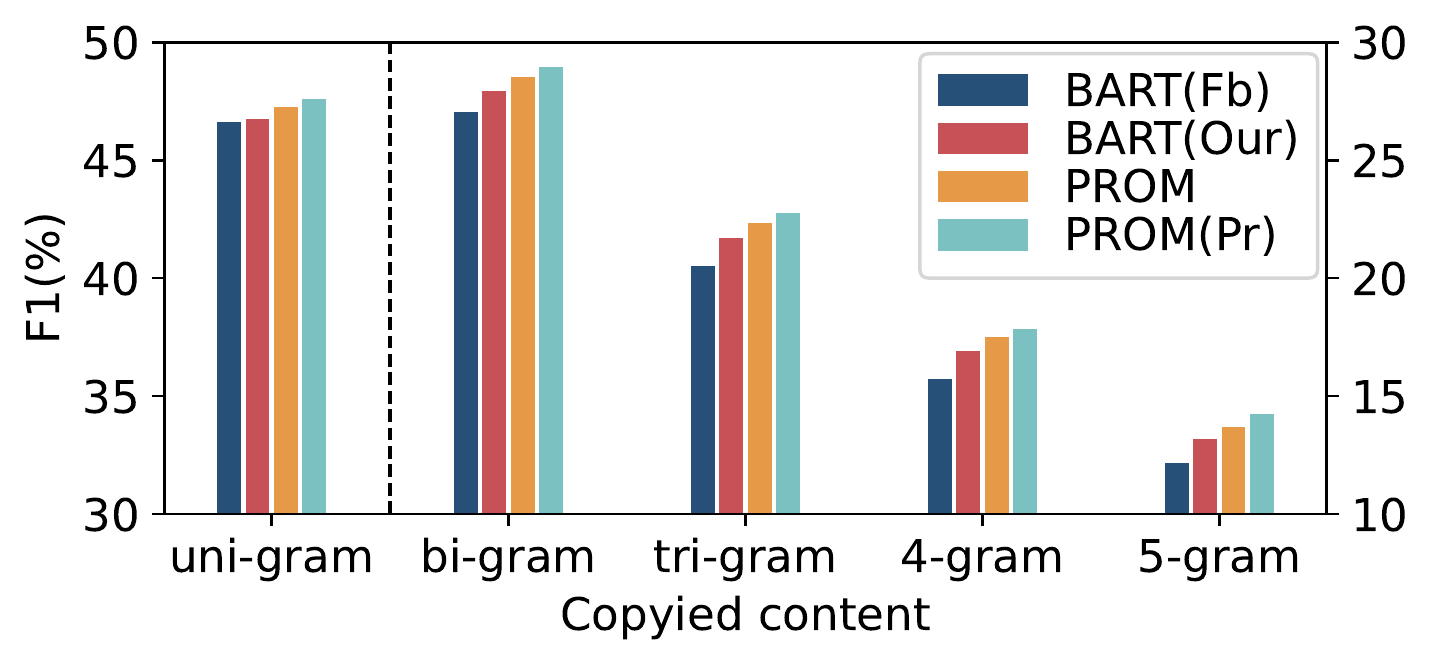}
		\caption{\label{content} $\operatorname{F}_1$ scores of copied $n$-grams on CNN/DM. ``\textit{PROM(Pr)}'' denotes results of PROM$_{pre-train}$.}
\end{figure}

To intuitively prove the effectiveness of PROM, we present an illustration of copied contents. We gather $n$-grams that exist in both the input and the reference (i.e. $n$-grams ought to be copied), and those that appear in both the input and the predicted summaries (i.e. actually copied), and then compute the $\operatorname{F}_1$ scores. As shown in Figure \ref{content}, PROM has advantages on each granularity but is better on bi-gram and longer. The pre-training makes continuous progress. This indicates that PROM corrects the copied contents compared to baselines.

\subsection{Faithfulness}\label{consis}
Recent work shows that PrLM-based abstractive summarization systems still suffer from unfaithfulness \cite{chen2022towards, wan-bansal-2022-factpegasus}. 
With the enhancement of copying, we discuss the faithfulness of our model in the fine-tuning setting.

\begin{table}[tbh]
	\centering\small
	{\begin{tabular}{p{2.8cm}p{0.9cm}p{0.9cm}p{0.9cm}}
		\toprule
            \multicolumn{4}{c}{\emph{Part 1: Factuality}} \\
            \midrule
		\textbf{Model} &\textbf{R-2} & \textbf{BERTS.} & \textbf{FactCC}  \\ 
            \midrule
            BART$_{large}$ $\dagger$  & 21.53 & 88.36 & 51.11  \\
            BART$_{large}$(Fb)  & 21.08 & 88.31 & 49.90 \\
            BART$_{large}$(Our) & 21.20 & 88.41 & 53.70 \\
            PEGASUS $\dagger$ & 21.47 & 88.27 &50.98  \\
            PROM & 21.59 & 88.46 & 57.39  \\
            \toprule
            \multicolumn{4}{c}{\emph{Part 2: Entity Coverage}} \\
            \midrule
            \textbf{Model}  & \textbf{p$_t$} &\textbf{r} &\textbf{F$_1$} \\ 
            \midrule
            BART$_{large}$(Fb)   & 65.39 & 74.32 & 69.57\\
            BART$_{large}$(Our)  & 67.29 & 71.89 & 69.52\\
            PROM   & 66.90 & 74.37 & 70.44 \\
            \toprule
            \multicolumn{4}{c}{\emph{Part 3: Human Evaluation}} \\
            \midrule
            \textbf{PROM}  & \textbf{Win(\%)} &\textbf{Tie(\%)} &\textbf{Lose(\%)} \\ 
            \midrule
            Faithfulness & 23.33 & 56.00 & 20.67 \\
            Informativenss   & 32.67 & 34.67 & 32.67 \\
            Readability   & 16.00 & 73.33 & 10.67 \\
		\bottomrule
	\end{tabular}
	}
        \caption{Advanced metrics. The results are from our models with PROM that are fine-tuned on CNN/DM. The three parts are described in Section \ref{consis} and \ref{human}. A dagger means that results are from previous work \citet{chen2022towards}.}
	\label{adm}
\end{table}

\subsubsection{Factuality metrics} \label{fac}
Following previous work \cite{chen2022towards, wan-bansal-2022-factpegasus, pagnoni-etal-2021-understanding}, we compute two factuality metrics that are highly correlated with human judgment.
(i) BERTScore \cite{bertscore} computes the similarity of contextual embeddings from BERT \cite{devlin-etal-2019-bert}.
(ii) FactCC \cite{factcc} is calculated by a weakly supervised model, which is trained to detect the consistencies and conflicts. 
The first part of Table \ref{adm} presents ROUGE-2 and factuality scores. The higher numbers of PROM show that besides more overlapped bi-grams, the copying method tends to produce more faithful and stable summaries.

\subsubsection{Entity Coverage} \label{en}
The consistency of named entities also embodies the faithfulness of a summary 
\cite{nan-etal-2021-entity, entity, chen2022towards}. 
Hence, we consider entity coverage and present precision, recall, and $\operatorname{F}_1$ between the predictions $Y_{prd}$ and the references $Y$. 
\begin{equation}
\setlength{\abovedisplayskip}{5pt}
\setlength{\belowdisplayskip}{5pt}
\begin{split}
&\operatorname{p}=\left|N E\left(Y\right) \cap N E\left(Y_{prd }\right)\right| /\left|N E\left(Y_{prd }\right)\right| \\
&\operatorname{r} =\left|N E\left(Y\right) \cap N E\left(Y_{prd }\right)\right| /\left|N E\left(Y\right)\right| \\
&\operatorname{F}_1 = 2 \cdot \operatorname{p} \cdot \operatorname{r} / (\operatorname{p} + \operatorname{r} )
\nonumber 
\end{split}
\label{prfen}
\end{equation}

The results are in the second part of Table \ref{adm}. Our model shows advantages on recall, little difference on precision, but still improves $\operatorname{F}_1$ scores. 
It suggests that our model generates more related entities, and contributes to faithfulness.

\subsection{Human Evaluation} \label{human}
As a complement to the above automatic metrics, we conduct human judgment to evaluate faithfulness, informativeness, and readability.
50 cases are randomly sampled and shuffled, then evaluated by 3 annotators \cite{wan-bansal-2022-factpegasus, cao-wang-2021-cliff}. Detailed implementation is shown in Appendix \ref{appc}.

The third part of Table \ref{adm} shows how PROM performs compared to the BART baseline. 
We can see that our model significantly wins BART in faithfulness (by 2.66\%) and readability (by 5.33\%) and ties in informativeness.

In summary, our method not only improves the predicted summaries in terms of the overlap with gold summaries, but also enhances the faithfulness compared to the source document, and better aligns with human preference.

\subsection{Large Language Models for Summarization}
Large language models have exhibited impressive capabilities in zero- and few-shot settings, leading to the new paradigm of prompting.
There are two application approaches of LLM, subscribing online API (like ChatGPT API \cite{ouyang2022training}) or running locally (like Llama \cite{touvron2023llama}). Both of them rely on computational resource support heavily, thus leaving positions for language models that are smaller but more expert.

To evaluate the zero-shot performance of the proposed method, those summarization tasks are tested on mainstream large language models for comparison. 
The implemented models are ChatGPT (gpt-3.5-turbo), Llama-2-7B \cite{touvron2023llama2}, Llama-2-13B, Llama-30B, and our model with PROM.
The prompt is ``\textit{Summarize the given document. Document: \{doc\} Summary:}''. The results of ROUGE are shown in Table \ref{llm}.
It is observed that our model surpasses Llama models in the zero-shot setting.
Compared to ChatGPT, our model achieves better scores on CNN/DM, NYT, and WikiHow, is comparable on BillSum and arXiv, and shows inferiority on Xsum. 

\begin{table}[tbh]
	\centering\small
        \setlength{\belowcaptionskip}{-0cm}
	{\begin{tabular}{p{1.7cm}p{0.4cm}p{0.4cm}p{0.5cm}p{0.4cm}p{0.4cm}p{0.4cm}}
		\toprule
		{\textbf{Model}} & \makebox[0.4cm][c]{\textbf{R$_{1}$}} & \textbf{R$_{2}$} & \textbf{R$_{L}$} & \textbf{R$_{1}$} & \textbf{R$_{2}$} & \textbf{R$_{L}$} \\ 
            \midrule
            & \multicolumn{3}{c}{\textbf{CNN/DM}} &\multicolumn{3}{c}{\textbf{XSum}}\\
		\midrule
            ChatGPT  &  \makebox[0.4cm][c]{30.81} & \makebox[0.4cm][c]{11.74} & \makebox[0.4cm][c]{28.54}  & \makebox[0.4cm][c]{25.48} & \makebox[0.4cm][c]{8.61} & \makebox[0.4cm][c]{21.54} \\ 
            Llama2-7B  &  \makebox[0.4cm][c]{23.89} & \makebox[0.4cm][c]{8.30} & \makebox[0.4cm][c]{21.83}  & \makebox[0.4cm][c]{17.30} & \makebox[0.4cm][c]{4.54} & \makebox[0.4cm][c]{14.41} \\ 
            
            \quad-13B  &  \makebox[0.4cm][c]{25.18} & \makebox[0.4cm][c]{9.03} & \makebox[0.4cm][c]{22.94}  & \makebox[0.4cm][c]{14.85} & \makebox[0.4cm][c]{2.53} & \makebox[0.4cm][c]{11.85} \\ 
            Llama-30B  &  \makebox[0.4cm][c]{25.82} & \makebox[0.4cm][c]{9.38} & \makebox[0.4cm][c]{23.40}  & \makebox[0.4cm][c]{15.16} & \makebox[0.4cm][c]{2.68} & \makebox[0.4cm][c]{12.11} \\ 
            
            PROM$_\textup{full}$   & \makebox[0.4cm][c]{37.87} &\makebox[0.4cm][c]{15.91} &\makebox[0.4cm][c]{34.16} & \makebox[0.4cm][c]{22.96} & \makebox[0.4cm][c]{6.05} & \makebox[0.4cm][c]{17.78} \\
            
		\midrule
            \multirow{2}{*}{} & \multicolumn{3}{c}{\textbf{NYT}} &\multicolumn{3}{c}{\textbf{WikiHow}} \\ 
            \midrule
            ChatGPT  &  \makebox[0.4cm][c]{32.31} & \makebox[0.4cm][c]{12.70} & \makebox[0.4cm][c]{28.03}  & \makebox[0.4cm][c]{21.10} & \makebox[0.4cm][c]{4.30} & \makebox[0.4cm][c]{19.86} \\ 
            Llama2-7B  &  \makebox[0.4cm][c]{25.03} & \makebox[0.4cm][c]{8.80} & \makebox[0.4cm][c]{20.90}  & \makebox[0.4cm][c]{20.49} & \makebox[0.4cm][c]{5.15} & \makebox[0.4cm][c]{19.53} \\ 
            \quad-13B  &  \makebox[0.4cm][c]{28.24} & \makebox[0.4cm][c]{10.95} & \makebox[0.4cm][c]{23.55}  & \makebox[0.4cm][c]{21.61} & \makebox[0.4cm][c]{4.95} & \makebox[0.4cm][c]{20.36} \\ 
            Llama-30B  &  \makebox[0.4cm][c]{28.44} & \makebox[0.4cm][c]{10.65} & \makebox[0.4cm][c]{23.66}  & \makebox[0.4cm][c]{22.20} & \makebox[0.4cm][c]{5.86} & \makebox[0.4cm][c]{20.93} \\ 
            
            PROM$_\textup{full}$  & \makebox[0.4cm][c]{36.95} &\makebox[0.4cm][c]{17.21} &\makebox[0.4cm][c]{29.42}  & \makebox[0.4cm][c]{25.90} & \makebox[0.4cm][c]{6.37} & \makebox[0.4cm][c]{24.08}\\
            \midrule
            \multirow{2}{*}{} & \multicolumn{3}{c}{\textbf{BillSum}} &\multicolumn{3}{c}{\textbf{arXiv}} \\ 
            \midrule
            
            ChatGPT  &  \makebox[0.4cm][c]{36.57} & \makebox[0.4cm][c]{19.09} & \makebox[0.4cm][c]{33.66}  & \makebox[0.4cm][c]{30.95} & \makebox[0.4cm][c]{10.74} & \makebox[0.4cm][c]{27.75} \\ 
            Llama2-7B  &  \makebox[0.4cm][c]{31.12} & \makebox[0.4cm][c]{15.21} & \makebox[0.4cm][c]{28.13}  & \makebox[0.4cm][c]{23.92} & \makebox[0.4cm][c]{7.65} & \makebox[0.4cm][c]{21.36} \\ 
            \quad-13B  &  \makebox[0.4cm][c]{32.14} & \makebox[0.4cm][c]{15.82} & \makebox[0.4cm][c]{29.07}  & \makebox[0.4cm][c]{27.80} & \makebox[0.4cm][c]{9.09} & \makebox[0.4cm][c]{24.61} \\ 
            Llama-30B  &  \makebox[0.4cm][c]{30.90} & \makebox[0.4cm][c]{14.62} & \makebox[0.4cm][c]{28.00}  & \makebox[0.4cm][c]{28.34} & \makebox[0.4cm][c]{8.44} & \makebox[0.4cm][c]{25.10} \\ 
            
            PROM$_\textup{full}$  & \makebox[0.4cm][c]{40.05} &\makebox[0.4cm][c]{14.66} &\makebox[0.4cm][c]{33.34}  & \makebox[0.4cm][c]{34.90} & \makebox[0.4cm][c]{9.71} & \makebox[0.4cm][c]{30.43} \\
            \bottomrule
	\end{tabular}
	}
        \caption{Zero-shot ROUGE $\operatorname{F}_1$ results of large language models. ChatGPT version is gpt-3.5-turbo.}
	\label{llm}
\end{table}

We also conduct human-like evaluations using G-EVAL \cite{liu2023gpteval}, which prompts ChatGPT to score the summaries. Zero-shot generations on CNN/DM and Xsum of PROM and Llama-13B are compared.\footnote{The prompt line follows \url{https://github.com/nlpyang/geval.}} We evaluated the summaries of 100 samples from PROM and Llama-13B. The average score on CNN/DM is 3.53 for PROPM and 3.36 for Llama-13B, and the score on Xsum is 3.57 for PROM and 3.02 for Llama-13B.
This suggests that continuous pre-training is an effective method for the zero-shot performance of a small expert model.
This also proves that zero-shot small models with our method show a certain degree of competitiveness in the age of LLM.

\subsection{Data Bias} \label{databias}
This section analyzes experimental results considering dataset features.
\vspace{3mm}

\noindent \textbf{Extractiveness \& Abstractiveness Level. }
As is shown in Section \ref{datasets}, downstream datasets vary in the tendency to extract fragments from the source document. In Table \ref{pretrain}, gains on BillSum and XSum are more moderate than other tasks. The possible reasons can be that (i) PROM breaks the continuous copying for phrase enhancement, which may disturb the extractions of BillSum. (ii) Xsum has a strong preference for abstraction and conciseness and thus benefits less from copying.
\vspace{3mm}

\begin{figure}[hbt]
		\centering
		\includegraphics[width=0.49\textwidth]{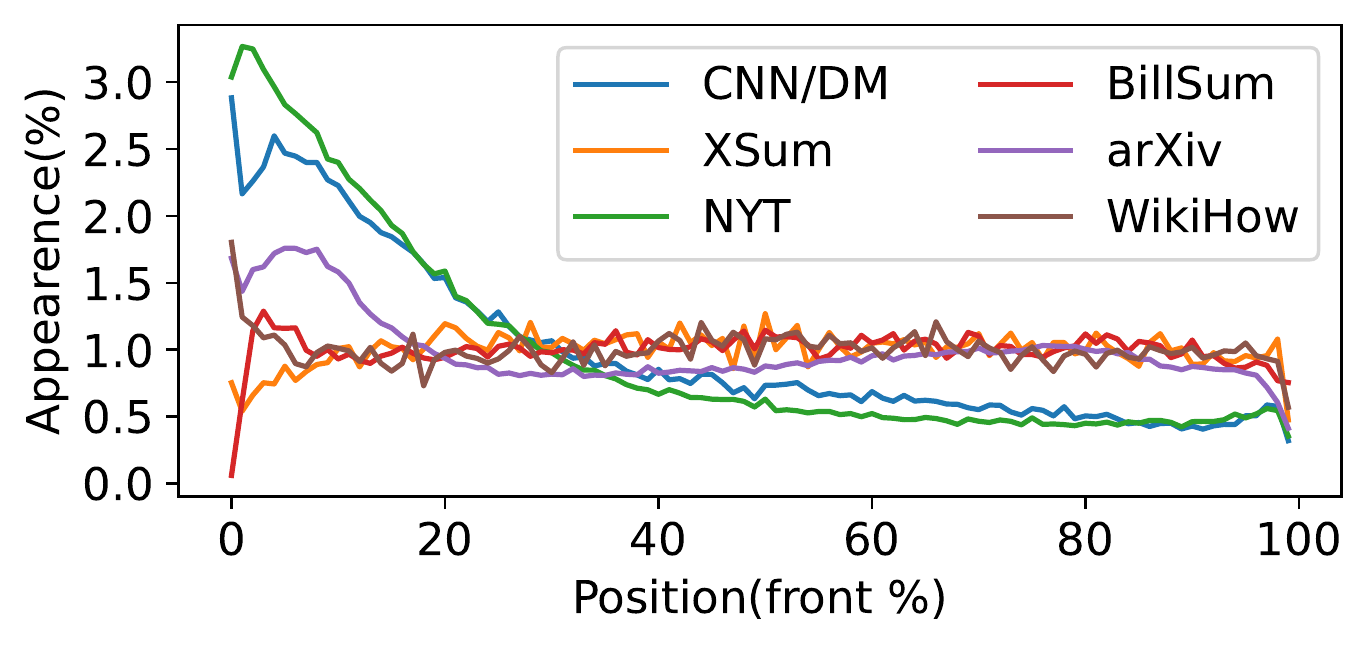}
		\caption{\label{pos} Position distributions of the overlaps across datasets. }
\end{figure}

\noindent \textbf{Lead Bias. } 
The position distribution of the salient information is another essential feature of summarization tasks. 
Lead bias can be leveraged to make large margin improvements in the unsupervised setting \cite{ted, wikitransfer}.
Figure \ref{pos} presents the position distributions of overlaps in the source documents. 
It shows that datasets have different levels of position preference. CNN/DM and NYT show stronger lead bias, while others have flatter curves.
\vspace{3mm}

Following \citet{ted} and \citet{wikitransfer}, we consider a loose unsupervised setting, where the model can not see summary data but can leverage this domain feature.
As mentioned in Section \ref{pt}, we create the lead-biased data using the 29G subset and implement continuous pre-training.
Zero-shot results in table \ref{lead} indicate that our method can achieve better scores with the lead bias. The results also prove that PROM can surpass TED and WikiTransfer methods under a similar setting and thus again verify the effectiveness.


\begin{table}[tbh]
	\centering\small
	{\begin{tabular}{p{3.4cm}p{0.6cm}p{0.6cm}p{0.6cm}}
		\toprule
            \textbf{Model} & \textbf{R$_{1}$} &\textbf{R$_{2}$} &\textbf{R$_{L}$}\\ 
		\midrule
            TED & 38.38 & 16.49 & 35.08 \\
            WikiTransfer (w/o bin) & 39.11 & 16.98 & 35.66 \\
            PROM  & 39.78 &17.19 &36.12 \\
		\bottomrule
	\end{tabular}
	}
        \caption{Zero-shot ROUGE $\operatorname{F}_1$ scores with lead bias on CNN/DM.}
	\label{lead}
\end{table}

\section{Conclusion}
In this work, we propose PROM, a novel method to enhance phrase copying, which makes contributions to abstractive summarization in both supervised and zero-shot settings.
This paper also gives a systematic study of copying methods for abstractive summarization.
The proposed PROM encourages the copying of phrases and surpasses previous copying methods in fine-tuning. 
PROM is further utilized in pre-training and achieves improvements in zero-shot performance on a wide range of summarization tasks.
The experimental results also prove that PROM shows advantages in faithfulness, entity coverage, and human evaluation. 
The scalability of PROM is shown across datasets that vary in genre, extractiveness level, and position bias. 
Our pre-trained model still has a practical significance in the zero-shot setting compared to large language models.

\bibliography{lrec-coling2024-example}

\begin{thebibliography}{76}
\expandafter\ifx\csname natexlab\endcsname\relax\def\natexlab#1{#1}\fi

\bibitem[{Bi et~al.(2020)Bi, Li, Wu, Yan, Wang, Huang, Huang, and Si}]{PALM}
Bin Bi, Chenliang Li, Chen Wu, Ming Yan, Wei Wang, Songfang Huang, Fei Huang, and Luo Si. 2020.
\newblock \href {https://doi.org/10.18653/v1/2020.emnlp-main.700} {{PALM:} pre-training an autoencoding{\&}autoregressive language model for context-conditioned generation}.
\newblock In \emph{Proceedings of the 2020 Conference on Empirical Methods in Natural Language Processing, {EMNLP} 2020, Online, November 16-20, 2020}, pages 8681--8691. Association for Computational Linguistics.

\bibitem[{Bra{\v{z}}inskas et~al.(2020)Bra{\v{z}}inskas, Lapata, and Titov}]{brazinskas-etal-2020-unsupervised}
Arthur Bra{\v{z}}inskas, Mirella Lapata, and Ivan Titov. 2020.
\newblock \href {https://doi.org/10.18653/v1/2020.acl-main.461} {Unsupervised opinion summarization as copycat-review generation}.
\newblock In \emph{Proceedings of the 58th Annual Meeting of the Association for Computational Linguistics}, pages 5151--5169, Online. Association for Computational Linguistics.

\bibitem[{Brazinskas et~al.(2020)Brazinskas, Lapata, and Titov}]{DBLP:conf/acl/BrazinskasLT20}
Arthur Brazinskas, Mirella Lapata, and Ivan Titov. 2020.
\newblock \href {https://doi.org/10.18653/v1/2020.acl-main.461} {Unsupervised opinion summarization as copycat-review generation}.
\newblock In \emph{Proceedings of the 58th Annual Meeting of the Association for Computational Linguistics, {ACL} 2020, Online, July 5-10, 2020}, pages 5151--5169. Association for Computational Linguistics.

\bibitem[{Brown et~al.(2020)Brown, Mann, Ryder, Subbiah, Kaplan, Dhariwal, Neelakantan, Shyam, Sastry, Askell et~al.}]{brown2020language}
Tom Brown, Benjamin Mann, Nick Ryder, Melanie Subbiah, Jared~D Kaplan, Prafulla Dhariwal, Arvind Neelakantan, Pranav Shyam, Girish Sastry, Amanda Askell, et~al. 2020.
\newblock Language models are few-shot learners.
\newblock \emph{Advances in neural information processing systems}, 33:1877--1901.

\bibitem[{Cao and Wang(2021)}]{cao-wang-2021-cliff}
Shuyang Cao and Lu~Wang. 2021.
\newblock \href {https://doi.org/10.18653/v1/2021.emnlp-main.532} {{CLIFF}: Contrastive learning for improving faithfulness and factuality in abstractive summarization}.
\newblock In \emph{Proceedings of the 2021 Conference on Empirical Methods in Natural Language Processing}, pages 6633--6649, Online and Punta Cana, Dominican Republic. Association for Computational Linguistics.

\bibitem[{Celikyilmaz et~al.(2018)Celikyilmaz, Bosselut, He, and Choi}]{celikyilmaz-etal-2018-deep}
Asli Celikyilmaz, Antoine Bosselut, Xiaodong He, and Yejin Choi. 2018.
\newblock \href {https://doi.org/10.18653/v1/N18-1150} {Deep communicating agents for abstractive summarization}.
\newblock In \emph{Proceedings of the 2018 Conference of the North {A}merican Chapter of the Association for Computational Linguistics: Human Language Technologies, Volume 1 (Long Papers)}, pages 1662--1675, New Orleans, Louisiana. Association for Computational Linguistics.

\bibitem[{Chen et~al.(2022)Chen, Li, Gao, and Zhang}]{chen2022towards}
Xiuying Chen, Mingzhe Li, Xin Gao, and Xiangliang Zhang. 2022.
\newblock Towards improving faithfulness in abstractive summarization.
\newblock \emph{arXiv preprint arXiv:2210.01877}.

\bibitem[{Chen et~al.(2020)Chen, Liu, Zhong, Dou, Wang, Qiu, and Huang}]{chen-etal-2020-cdevalsumm}
Yiran Chen, Pengfei Liu, Ming Zhong, Zi-Yi Dou, Danqing Wang, Xipeng Qiu, and Xuanjing Huang. 2020.
\newblock \href {https://doi.org/10.18653/v1/2020.findings-emnlp.329} {{CDE}val{S}umm: An empirical study of cross-dataset evaluation for neural summarization systems}.
\newblock In \emph{Findings of the Association for Computational Linguistics: EMNLP 2020}, pages 3679--3691, Online. Association for Computational Linguistics.

\bibitem[{Chowdhery et~al.(2022)Chowdhery, Narang, Devlin, Bosma, Mishra, Roberts, Barham, Chung, Sutton, Gehrmann et~al.}]{chowdhery2022palm}
Aakanksha Chowdhery, Sharan Narang, Jacob Devlin, Maarten Bosma, Gaurav Mishra, Adam Roberts, Paul Barham, Hyung~Won Chung, Charles Sutton, Sebastian Gehrmann, et~al. 2022.
\newblock Palm: Scaling language modeling with pathways.
\newblock \emph{arXiv preprint arXiv:2204.02311}.

\bibitem[{Cohan et~al.(2018)Cohan, Dernoncourt, Kim, Bui, Kim, Chang, and Goharian}]{arxiv}
Arman Cohan, Franck Dernoncourt, Doo~Soon Kim, Trung Bui, Seokhwan Kim, Walter Chang, and Nazli Goharian. 2018.
\newblock \href {https://doi.org/10.18653/v1/N18-2097} {A discourse-aware attention model for abstractive summarization of long documents}.
\newblock In \emph{Proceedings of the 2018 Conference of the North {A}merican Chapter of the Association for Computational Linguistics: Human Language Technologies, Volume 2 (Short Papers)}, pages 615--621, New Orleans, Louisiana. Association for Computational Linguistics.

\bibitem[{Devlin et~al.(2019)Devlin, Chang, Lee, and Toutanova}]{devlin-etal-2019-bert}
Jacob Devlin, Ming-Wei Chang, Kenton Lee, and Kristina Toutanova. 2019.
\newblock \href {https://doi.org/10.18653/v1/N19-1423} {{BERT}: Pre-training of deep bidirectional transformers for language understanding}.
\newblock In \emph{Proceedings of the 2019 Conference of the North {A}merican Chapter of the Association for Computational Linguistics: Human Language Technologies, Volume 1 (Long and Short Papers)}, pages 4171--4186, Minneapolis, Minnesota. Association for Computational Linguistics.

\bibitem[{Dinan et~al.(2019)Dinan, Roller, Shuster, Fan, Auli, and Weston}]{DBLP:conf/iclr/DinanRSFAW19}
Emily Dinan, Stephen Roller, Kurt Shuster, Angela Fan, Michael Auli, and Jason Weston. 2019.
\newblock \href {https://openreview.net/forum?id=r1l73iRqKm} {Wizard of wikipedia: Knowledge-powered conversational agents}.
\newblock In \emph{7th International Conference on Learning Representations, {ICLR} 2019, New Orleans, LA, USA, May 6-9, 2019}. OpenReview.net.

\bibitem[{Dong et~al.(2019)Dong, Yang, Wang, Wei, Liu, Wang, Gao, Zhou, and Hon}]{DBLP:conf/nips/00040WWLWGZH19}
Li~Dong, Nan Yang, Wenhui Wang, Furu Wei, Xiaodong Liu, Yu~Wang, Jianfeng Gao, Ming Zhou, and Hsiao{-}Wuen Hon. 2019.
\newblock \href {https://proceedings.neurips.cc/paper/2019/hash/c20bb2d9a50d5ac1f713f8b34d9aac5a-Abstract.html} {Unified language model pre-training for natural language understanding and generation}.
\newblock In \emph{Advances in Neural Information Processing Systems 32: Annual Conference on Neural Information Processing Systems 2019, NeurIPS 2019, December 8-14, 2019, Vancouver, BC, Canada}, pages 13042--13054.

\bibitem[{Dou et~al.(2021)Dou, Liu, Hayashi, Jiang, and Neubig}]{dou-etal-2021-gsum}
Zi-Yi Dou, Pengfei Liu, Hiroaki Hayashi, Zhengbao Jiang, and Graham Neubig. 2021.
\newblock \href {https://doi.org/10.18653/v1/2021.naacl-main.384} {{GS}um: A general framework for guided neural abstractive summarization}.
\newblock In \emph{Proceedings of the 2021 Conference of the North American Chapter of the Association for Computational Linguistics: Human Language Technologies}, pages 4830--4842, Online. Association for Computational Linguistics.

\bibitem[{Edunov et~al.(2019)Edunov, Baevski, and Auli}]{edunov-etal-2019-pre}
Sergey Edunov, Alexei Baevski, and Michael Auli. 2019.
\newblock \href {https://doi.org/10.18653/v1/N19-1409} {Pre-trained language model representations for language generation}.
\newblock In \emph{Proceedings of the 2019 Conference of the North {A}merican Chapter of the Association for Computational Linguistics: Human Language Technologies, Volume 1 (Long and Short Papers)}, pages 4052--4059, Minneapolis, Minnesota. Association for Computational Linguistics.

\bibitem[{Fabbri et~al.(2021{\natexlab{a}})Fabbri, Han, Li, Li, Ghazvininejad, Joty, Radev, and Mehdad}]{wikitransfer}
Alexander Fabbri, Simeng Han, Haoyuan Li, Haoran Li, Marjan Ghazvininejad, Shafiq Joty, Dragomir Radev, and Yashar Mehdad. 2021{\natexlab{a}}.
\newblock \href {https://doi.org/10.18653/v1/2021.naacl-main.57} {Improving zero and few-shot abstractive summarization with intermediate fine-tuning and data augmentation}.
\newblock In \emph{Proceedings of the 2021 Conference of the North American Chapter of the Association for Computational Linguistics: Human Language Technologies}, pages 704--717, Online. Association for Computational Linguistics.

\bibitem[{Fabbri et~al.(2021{\natexlab{b}})Fabbri, Han, Li, Li, Ghazvininejad, Joty, Radev, and Mehdad}]{DBLP:conf/naacl/FabbriHLLGJRM21}
Alexander~R. Fabbri, Simeng Han, Haoyuan Li, Haoran Li, Marjan Ghazvininejad, Shafiq~R. Joty, Dragomir~R. Radev, and Yashar Mehdad. 2021{\natexlab{b}}.
\newblock \href {https://doi.org/10.18653/v1/2021.naacl-main.57} {Improving zero and few-shot abstractive summarization with intermediate fine-tuning and data augmentation}.
\newblock In \emph{Proceedings of the 2021 Conference of the North American Chapter of the Association for Computational Linguistics: Human Language Technologies, {NAACL-HLT} 2021, Online, June 6-11, 2021}, pages 704--717. Association for Computational Linguistics.

\bibitem[{Gehrmann et~al.(2018)Gehrmann, Deng, and Rush}]{bottomup}
Sebastian Gehrmann, Yuntian Deng, and Alexander~M. Rush. 2018.
\newblock \href {https://doi.org/10.18653/v1/d18-1443} {Bottom-up abstractive summarization}.
\newblock In \emph{Proceedings of the 2018 Conference on Empirical Methods in Natural Language Processing, Brussels, Belgium, October 31 - November 4, 2018}, pages 4098--4109. Association for Computational Linguistics.

\bibitem[{Gliwa et~al.(2019)Gliwa, Mochol, Biesek, and Wawer}]{samsum}
Bogdan Gliwa, Iwona Mochol, Maciej Biesek, and Aleksander Wawer. 2019.
\newblock \href {http://arxiv.org/abs/1911.12237} {Samsum corpus: {A} human-annotated dialogue dataset for abstractive summarization}.
\newblock \emph{CoRR}, abs/1911.12237.

\bibitem[{Grusky et~al.(2018)Grusky, Naaman, and Artzi}]{grusky-etal-2018-newsroom}
Max Grusky, Mor Naaman, and Yoav Artzi. 2018.
\newblock \href {https://doi.org/10.18653/v1/N18-1065} {{N}ewsroom: A dataset of 1.3 million summaries with diverse extractive strategies}.
\newblock In \emph{Proceedings of the 2018 Conference of the North {A}merican Chapter of the Association for Computational Linguistics: Human Language Technologies, Volume 1 (Long Papers)}, pages 708--719, New Orleans, Louisiana. Association for Computational Linguistics.

\bibitem[{Gu et~al.(2016)Gu, Lu, Li, and Li}]{copynet}
Jiatao Gu, Zhengdong Lu, Hang Li, and Victor O.~K. Li. 2016.
\newblock \href {https://doi.org/10.18653/v1/p16-1154} {Incorporating copying mechanism in sequence-to-sequence learning}.
\newblock In \emph{Proceedings of the 54th Annual Meeting of the Association for Computational Linguistics, {ACL} 2016, August 7-12, 2016, Berlin, Germany, Volume 1: Long Papers}. The Association for Computer Linguistics.

\bibitem[{Gururangan et~al.(2020)Gururangan, Marasovi{\'c}, Swayamdipta, Lo, Beltagy, Downey, and Smith}]{gururangan-etal-2020-dont}
Suchin Gururangan, Ana Marasovi{\'c}, Swabha Swayamdipta, Kyle Lo, Iz~Beltagy, Doug Downey, and Noah~A. Smith. 2020.
\newblock \href {https://doi.org/10.18653/v1/2020.acl-main.740} {Don{'}t stop pretraining: Adapt language models to domains and tasks}.
\newblock In \emph{Proceedings of the 58th Annual Meeting of the Association for Computational Linguistics}, pages 8342--8360, Online. Association for Computational Linguistics.

\bibitem[{He et~al.(2022)He, Peng, Lu, Wang, Mei, Liu, Xu, Awadalla, Shi, Zhu, Xiong, Zeng, Gao, and Huang}]{DBLP:journals/corr/abs-2208-09770}
Pengcheng He, Baolin Peng, Liyang Lu, Song Wang, Jie Mei, Yang Liu, Ruochen Xu, Hany~Hassan Awadalla, Yu~Shi, Chenguang Zhu, Wayne Xiong, Michael Zeng, Jianfeng Gao, and Xuedong Huang. 2022.
\newblock \href {https://doi.org/10.48550/arXiv.2208.09770} {Z-code++: {A} pre-trained language model optimized for abstractive summarization}.
\newblock \emph{CoRR}, abs/2208.09770.

\bibitem[{Hu et~al.(2015)Hu, Chen, and Zhu}]{LCSTS}
Baotian Hu, Qingcai Chen, and Fangze Zhu. 2015.
\newblock \href {https://doi.org/10.18653/v1/d15-1229} {{LCSTS:} {A} large scale chinese short text summarization dataset}.
\newblock In \emph{Proceedings of the 2015 Conference on Empirical Methods in Natural Language Processing, {EMNLP} 2015, Lisbon, Portugal, September 17-21, 2015}, pages 1967--1972. The Association for Computational Linguistics.

\bibitem[{Hua and Wang(2017)}]{DBLP:conf/emnlp/HuaW17}
Xinyu Hua and Lu~Wang. 2017.
\newblock \href {https://doi.org/10.18653/v1/w17-4513} {A pilot study of domain adaptation effect for neural abstractive summarization}.
\newblock In \emph{Proceedings of the Workshop on New Frontiers in Summarization, NFiS@EMNLP 2017, Copenhagen, Denmark, September 7, 2017}, pages 100--106. Association for Computational Linguistics.

\bibitem[{Kim et~al.(2019)Kim, Kim, and Kim}]{DBLP:conf/naacl/KimKK19}
Byeongchang Kim, Hyunwoo Kim, and Gunhee Kim. 2019.
\newblock \href {https://doi.org/10.18653/v1/n19-1260} {Abstractive summarization of reddit posts with multi-level memory networks}.
\newblock In \emph{Proceedings of the 2019 Conference of the North American Chapter of the Association for Computational Linguistics: Human Language Technologies, {NAACL-HLT} 2019, Minneapolis, MN, USA, June 2-7, 2019, Volume 1 (Long and Short Papers)}, pages 2519--2531. Association for Computational Linguistics.

\bibitem[{Kornilova and Eidelman(2019)}]{kornilova-eidelman-2019-billsum}
Anastassia Kornilova and Vladimir Eidelman. 2019.
\newblock \href {https://doi.org/10.18653/v1/D19-5406} {{B}ill{S}um: A corpus for automatic summarization of {US} legislation}.
\newblock In \emph{Proceedings of the 2nd Workshop on New Frontiers in Summarization}, pages 48--56, Hong Kong, China. Association for Computational Linguistics.

\bibitem[{Koupaee and Wang(2018)}]{wikihow}
Mahnaz Koupaee and William~Yang Wang. 2018.
\newblock \href {http://arxiv.org/abs/1810.09305} {Wikihow: {A} large scale text summarization dataset}.
\newblock \emph{CoRR}, abs/1810.09305.

\bibitem[{Kryscinski et~al.(2020{\natexlab{a}})Kryscinski, McCann, Xiong, and Socher}]{kryscinski-etal-2020-evaluating}
Wojciech Kryscinski, Bryan McCann, Caiming Xiong, and Richard Socher. 2020{\natexlab{a}}.
\newblock \href {https://doi.org/10.18653/v1/2020.emnlp-main.750} {Evaluating the factual consistency of abstractive text summarization}.
\newblock In \emph{Proceedings of the 2020 Conference on Empirical Methods in Natural Language Processing (EMNLP)}, pages 9332--9346, Online. Association for Computational Linguistics.

\bibitem[{Kryscinski et~al.(2020{\natexlab{b}})Kryscinski, McCann, Xiong, and Socher}]{DBLP:conf/emnlp/KryscinskiMXS20}
Wojciech Kryscinski, Bryan McCann, Caiming Xiong, and Richard Socher. 2020{\natexlab{b}}.
\newblock \href {https://doi.org/10.18653/v1/2020.emnlp-main.750} {Evaluating the factual consistency of abstractive text summarization}.
\newblock In \emph{Proceedings of the 2020 Conference on Empirical Methods in Natural Language Processing, {EMNLP} 2020, Online, November 16-20, 2020}, pages 9332--9346. Association for Computational Linguistics.

\bibitem[{Kryscinski et~al.(2020{\natexlab{c}})Kryscinski, McCann, Xiong, and Socher}]{factcc}
Wojciech Kryscinski, Bryan McCann, Caiming Xiong, and Richard Socher. 2020{\natexlab{c}}.
\newblock \href {https://doi.org/10.18653/v1/2020.emnlp-main.750} {Evaluating the factual consistency of abstractive text summarization}.
\newblock In \emph{Proceedings of the 2020 Conference on Empirical Methods in Natural Language Processing, {EMNLP} 2020, Online, November 16-20, 2020}, pages 9332--9346. Association for Computational Linguistics.

\bibitem[{Lewis et~al.(2020)Lewis, Liu, Goyal, Ghazvininejad, Mohamed, Levy, Stoyanov, and Zettlemoyer}]{bart}
Mike Lewis, Yinhan Liu, Naman Goyal, Marjan Ghazvininejad, Abdelrahman Mohamed, Omer Levy, Veselin Stoyanov, and Luke Zettlemoyer. 2020.
\newblock \href {https://doi.org/10.18653/v1/2020.acl-main.703} {{BART:} denoising sequence-to-sequence pre-training for natural language generation, translation, and comprehension}.
\newblock In \emph{Proceedings of the 58th Annual Meeting of the Association for Computational Linguistics, {ACL} 2020, Online, July 5-10, 2020}, pages 7871--7880. Association for Computational Linguistics.

\bibitem[{Li et~al.(2021)Li, Xu, Yuan, Wang, Wu, He, and Zhou}]{coconet}
Haoran Li, Song Xu, Peng Yuan, Yujia Wang, Youzheng Wu, Xiaodong He, and Bowen Zhou. 2021.
\newblock \href {https://doi.org/10.18653/v1/2021.emnlp-main.336} {Learn to copy from the copying history: Correlational copy network for abstractive summarization}.
\newblock In \emph{Proceedings of the 2021 Conference on Empirical Methods in Natural Language Processing, {EMNLP} 2021, Virtual Event / Punta Cana, Dominican Republic, 7-11 November, 2021}, pages 4091--4101. Association for Computational Linguistics.

\bibitem[{Liu et~al.(2022{\natexlab{a}})Liu, Gao, Bai, Li, Hu, Huang, and Chen}]{DBLP:conf/coling/LiuGBLHHC22}
Xiaochen Liu, Yang Gao, Yu~Bai, Jiawei Li, Yinan Hu, Heyan Huang, and Boxing Chen. 2022{\natexlab{a}}.
\newblock \href {https://aclanthology.org/2022.coling-1.553} {{PSP:} pre-trained soft prompts for few-shot abstractive summarization}.
\newblock In \emph{Proceedings of the 29th International Conference on Computational Linguistics, {COLING} 2022, Gyeongju, Republic of Korea, October 12-17, 2022}, pages 6355--6368. International Committee on Computational Linguistics.

\bibitem[{Liu et~al.(2023)Liu, Iter, Xu, Wang, Xu, and Zhu}]{liu2023gpteval}
Yang Liu, Dan Iter, Yichong Xu, Shuohang Wang, Ruochen Xu, and Chenguang Zhu. 2023.
\newblock Gpteval: Nlg evaluation using gpt-4 with better human alignment.
\newblock \emph{arXiv preprint arXiv:2303.16634}.

\bibitem[{Liu and Lapata(2019)}]{liu-lapata-2019-text}
Yang Liu and Mirella Lapata. 2019.
\newblock \href {https://doi.org/10.18653/v1/D19-1387} {Text summarization with pretrained encoders}.
\newblock In \emph{Proceedings of the 2019 Conference on Empirical Methods in Natural Language Processing and the 9th International Joint Conference on Natural Language Processing (EMNLP-IJCNLP)}, pages 3730--3740, Hong Kong, China. Association for Computational Linguistics.

\bibitem[{Liu et~al.(2022{\natexlab{b}})Liu, Liu, Radev, and Neubig}]{liu-etal-2022-brio}
Yixin Liu, Pengfei Liu, Dragomir Radev, and Graham Neubig. 2022{\natexlab{b}}.
\newblock \href {https://doi.org/10.18653/v1/2022.acl-long.207} {{BRIO}: Bringing order to abstractive summarization}.
\newblock In \emph{Proceedings of the 60th Annual Meeting of the Association for Computational Linguistics (Volume 1: Long Papers)}, pages 2890--2903, Dublin, Ireland. Association for Computational Linguistics.

\bibitem[{Nallapati et~al.(2017)Nallapati, Zhai, and Zhou}]{nallapati2017summarunner}
Ramesh Nallapati, Feifei Zhai, and Bowen Zhou. 2017.
\newblock Summarunner: A recurrent neural network based sequence model for extractive summarization of documents.
\newblock In \emph{Thirty-first AAAI conference on artificial intelligence}.

\bibitem[{Nallapati et~al.(2016{\natexlab{a}})Nallapati, Zhou, dos Santos, Gu̇l{\c{c}}ehre, and Xiang}]{nallapati-etal-2016-cnndm}
Ramesh Nallapati, Bowen Zhou, Cicero dos Santos, {\c{C}}a{\u{g}}lar Gu̇l{\c{c}}ehre, and Bing Xiang. 2016{\natexlab{a}}.
\newblock \href {https://doi.org/10.18653/v1/K16-1028} {Abstractive text summarization using sequence-to-sequence {RNN}s and beyond}.
\newblock In \emph{Proceedings of the 20th {SIGNLL} Conference on Computational Natural Language Learning}, pages 280--290, Berlin, Germany. Association for Computational Linguistics.

\bibitem[{Nallapati et~al.(2016{\natexlab{b}})Nallapati, Zhou, dos Santos, G{\"{u}}l{\c{c}}ehre, and Xiang}]{DBLP:conf/conll/NallapatiZSGX16}
Ramesh Nallapati, Bowen Zhou, C{\'{\i}}cero~Nogueira dos Santos, {\c{C}}aglar G{\"{u}}l{\c{c}}ehre, and Bing Xiang. 2016{\natexlab{b}}.
\newblock \href {https://doi.org/10.18653/v1/k16-1028} {Abstractive text summarization using sequence-to-sequence rnns and beyond}.
\newblock In \emph{Proceedings of the 20th {SIGNLL} Conference on Computational Natural Language Learning, CoNLL 2016, Berlin, Germany, August 11-12, 2016}, pages 280--290. {ACL}.

\bibitem[{Nan et~al.(2021)Nan, Nallapati, Wang, Nogueira~dos Santos, Zhu, Zhang, McKeown, and Xiang}]{nan-etal-2021-entity}
Feng Nan, Ramesh Nallapati, Zhiguo Wang, Cicero Nogueira~dos Santos, Henghui Zhu, Dejiao Zhang, Kathleen McKeown, and Bing Xiang. 2021.
\newblock \href {https://doi.org/10.18653/v1/2021.eacl-main.235} {Entity-level factual consistency of abstractive text summarization}.
\newblock In \emph{Proceedings of the 16th Conference of the European Chapter of the Association for Computational Linguistics: Main Volume}, pages 2727--2733, Online. Association for Computational Linguistics.

\bibitem[{Narayan et~al.(2018)Narayan, Cohen, and Lapata}]{xsum}
Shashi Narayan, Shay~B. Cohen, and Mirella Lapata. 2018.
\newblock \href {https://doi.org/10.18653/v1/D18-1206} {Don{'}t give me the details, just the summary! topic-aware convolutional neural networks for extreme summarization}.
\newblock In \emph{Proceedings of the 2018 Conference on Empirical Methods in Natural Language Processing}, pages 1797--1807, Brussels, Belgium. Association for Computational Linguistics.

\bibitem[{Ouyang et~al.(2022)Ouyang, Wu, Jiang, Almeida, Wainwright, Mishkin, Zhang, Agarwal, Slama, Ray et~al.}]{ouyang2022training}
Long Ouyang, Jeffrey Wu, Xu~Jiang, Diogo Almeida, Carroll Wainwright, Pamela Mishkin, Chong Zhang, Sandhini Agarwal, Katarina Slama, Alex Ray, et~al. 2022.
\newblock Training language models to follow instructions with human feedback.
\newblock \emph{Advances in Neural Information Processing Systems}, 35:27730--27744.

\bibitem[{Pagnoni et~al.(2021)Pagnoni, Balachandran, and Tsvetkov}]{pagnoni-etal-2021-understanding}
Artidoro Pagnoni, Vidhisha Balachandran, and Yulia Tsvetkov. 2021.
\newblock \href {https://doi.org/10.18653/v1/2021.naacl-main.383} {Understanding factuality in abstractive summarization with {FRANK}: A benchmark for factuality metrics}.
\newblock In \emph{Proceedings of the 2021 Conference of the North American Chapter of the Association for Computational Linguistics: Human Language Technologies}, pages 4812--4829, Online. Association for Computational Linguistics.

\bibitem[{Paulus et~al.(2018)Paulus, Xiong, and Socher}]{DBLP:conf/iclr/PaulusXS18}
Romain Paulus, Caiming Xiong, and Richard Socher. 2018.
\newblock \href {https://openreview.net/forum?id=HkAClQgA-} {A deep reinforced model for abstractive summarization}.
\newblock In \emph{6th International Conference on Learning Representations, {ICLR} 2018, Vancouver, BC, Canada, April 30 - May 3, 2018, Conference Track Proceedings}. OpenReview.net.

\bibitem[{Qi et~al.(2020{\natexlab{a}})Qi, Yan, Gong, Liu, Duan, Chen, Zhang, and Zhou}]{qi2020prophetnet}
Weizhen Qi, Yu~Yan, Yeyun Gong, Dayiheng Liu, Nan Duan, Jiusheng Chen, Ruofei Zhang, and Ming Zhou. 2020{\natexlab{a}}.
\newblock Prophetnet: Predicting future n-gram for sequence-to-sequence pre-training.
\newblock In \emph{Proceedings of the 2020 Conference on Empirical Methods in Natural Language Processing: Findings}, pages 2401--2410.

\bibitem[{Qi et~al.(2020{\natexlab{b}})Qi, Yan, Gong, Liu, Duan, Chen, Zhang, and Zhou}]{DBLP:conf/emnlp/QiYGLDCZ020}
Weizhen Qi, Yu~Yan, Yeyun Gong, Dayiheng Liu, Nan Duan, Jiusheng Chen, Ruofei Zhang, and Ming Zhou. 2020{\natexlab{b}}.
\newblock \href {https://doi.org/10.18653/v1/2020.findings-emnlp.217} {Prophetnet: Predicting future n-gram for sequence-to-sequence pre-training}.
\newblock In \emph{Findings of the Association for Computational Linguistics: {EMNLP} 2020, Online Event, 16-20 November 2020}, volume {EMNLP} 2020 of \emph{Findings of {ACL}}, pages 2401--2410. Association for Computational Linguistics.

\bibitem[{Raffel et~al.(2020{\natexlab{a}})Raffel, Shazeer, Roberts, Lee, Narang, Matena, Zhou, Li, and Liu}]{2020t5}
Colin Raffel, Noam Shazeer, Adam Roberts, Katherine Lee, Sharan Narang, Michael Matena, Yanqi Zhou, Wei Li, and Peter~J. Liu. 2020{\natexlab{a}}.
\newblock \href {http://jmlr.org/papers/v21/20-074.html} {Exploring the limits of transfer learning with a unified text-to-text transformer}.
\newblock \emph{Journal of Machine Learning Research}, 21(140):1--67.

\bibitem[{Raffel et~al.(2020{\natexlab{b}})Raffel, Shazeer, Roberts, Lee, Narang, Matena, Zhou, Li, and Liu}]{DBLP:journals/jmlr/RaffelSRLNMZLL20}
Colin Raffel, Noam Shazeer, Adam Roberts, Katherine Lee, Sharan Narang, Michael Matena, Yanqi Zhou, Wei Li, and Peter~J. Liu. 2020{\natexlab{b}}.
\newblock \href {http://jmlr.org/papers/v21/20-074.html} {Exploring the limits of transfer learning with a unified text-to-text transformer}.
\newblock \emph{J. Mach. Learn. Res.}, 21:140:1--140:67.

\bibitem[{Rashkin et~al.(2019)Rashkin, Smith, Li, and Boureau}]{DBLP:conf/acl/RashkinSLB19}
Hannah Rashkin, Eric~Michael Smith, Margaret Li, and Y{-}Lan Boureau. 2019.
\newblock \href {https://doi.org/10.18653/v1/p19-1534} {Towards empathetic open-domain conversation models: {A} new benchmark and dataset}.
\newblock In \emph{Proceedings of the 57th Conference of the Association for Computational Linguistics, {ACL} 2019, Florence, Italy, July 28- August 2, 2019, Volume 1: Long Papers}, pages 5370--5381. Association for Computational Linguistics.

\bibitem[{Rush et~al.(2015)Rush, Chopra, and Weston}]{rush-etal-2015-neural}
Alexander~M. Rush, Sumit Chopra, and Jason Weston. 2015.
\newblock \href {https://doi.org/10.18653/v1/D15-1044} {A neural attention model for abstractive sentence summarization}.
\newblock In \emph{Proceedings of the 2015 Conference on Empirical Methods in Natural Language Processing}, pages 379--389, Lisbon, Portugal. Association for Computational Linguistics.

\bibitem[{Saggion and Poibeau(2013)}]{saggion2013automatic}
Horacio Saggion and Thierry Poibeau. 2013.
\newblock Automatic text summarization: Past, present and future.
\newblock In \emph{Multi-source, multilingual information extraction and summarization}, pages 3--21. Springer.

\bibitem[{See et~al.(2017)See, Liu, and Manning}]{pointer}
Abigail See, Peter~J. Liu, and Christopher~D. Manning. 2017.
\newblock \href {https://doi.org/10.18653/v1/P17-1099} {Get to the point: Summarization with pointer-generator networks}.
\newblock In \emph{Proceedings of the 55th Annual Meeting of the Association for Computational Linguistics, {ACL} 2017, Vancouver, Canada, July 30 - August 4, Volume 1: Long Papers}, pages 1073--1083. Association for Computational Linguistics.

\bibitem[{Sharma et~al.(2019)Sharma, Li, and Wang}]{sharma-etal-2019-bigpatent}
Eva Sharma, Chen Li, and Lu~Wang. 2019.
\newblock \href {https://doi.org/10.18653/v1/P19-1212} {{BIGPATENT}: A large-scale dataset for abstractive and coherent summarization}.
\newblock In \emph{Proceedings of the 57th Annual Meeting of the Association for Computational Linguistics}, pages 2204--2213, Florence, Italy. Association for Computational Linguistics.

\bibitem[{Song et~al.(2019)Song, Tan, Qin, Lu, and Liu}]{song2019mass}
Kaitao Song, Xu~Tan, Tao Qin, Jianfeng Lu, and Tie-Yan Liu. 2019.
\newblock Mass: Masked sequence to sequence pre-training for language generation.
\newblock \emph{arXiv preprint arXiv:1905.02450}.

\bibitem[{Touvron et~al.(2023{\natexlab{a}})Touvron, Lavril, Izacard, Martinet, Lachaux, Lacroix, Rozi{\`e}re, Goyal, Hambro, Azhar, Rodriguez, Joulin, Grave, and Lample}]{touvron2023llama}
Hugo Touvron, Thibaut Lavril, Gautier Izacard, Xavier Martinet, Marie-Anne Lachaux, Timoth{\'e}e Lacroix, Baptiste Rozi{\`e}re, Naman Goyal, Eric Hambro, Faisal Azhar, Aurelien Rodriguez, Armand Joulin, Edouard Grave, and Guillaume Lample. 2023{\natexlab{a}}.
\newblock \href {https://arxiv.org/abs/2302.13971} {Llama: Open and efficient foundation language models}.
\newblock \emph{ArXiv preprint}, abs/2302.13971.

\bibitem[{Touvron et~al.(2023{\natexlab{b}})Touvron, Martin, Stone, Albert, Almahairi, Babaei, Bashlykov, Batra, Bhargava, Bhosale et~al.}]{touvron2023llama2}
Hugo Touvron, Louis Martin, Kevin Stone, Peter Albert, Amjad Almahairi, Yasmine Babaei, Nikolay Bashlykov, Soumya Batra, Prajjwal Bhargava, Shruti Bhosale, et~al. 2023{\natexlab{b}}.
\newblock \href {https://arxiv.org/abs/2307.09288} {Llama 2: Open foundation and fine-tuned chat models}.
\newblock \emph{ArXiv preprint}, abs/2307.09288.

\bibitem[{Vaswani et~al.(2017)Vaswani, Shazeer, Parmar, Uszkoreit, Jones, Gomez, Kaiser, and Polosukhin}]{vaswani2017attention}
Ashish Vaswani, Noam Shazeer, Niki Parmar, Jakob Uszkoreit, Llion Jones, Aidan~N Gomez, {\L}ukasz Kaiser, and Illia Polosukhin. 2017.
\newblock Attention is all you need.
\newblock \emph{Advances in neural information processing systems}, 30.

\bibitem[{Wan and Bansal(2022)}]{wan-bansal-2022-factpegasus}
David Wan and Mohit Bansal. 2022.
\newblock \href {https://doi.org/10.18653/v1/2022.naacl-main.74} {{F}act{PEGASUS}: Factuality-aware pre-training and fine-tuning for abstractive summarization}.
\newblock In \emph{Proceedings of the 2022 Conference of the North American Chapter of the Association for Computational Linguistics: Human Language Technologies}, pages 1010--1028, Seattle, United States. Association for Computational Linguistics.

\bibitem[{Wang et~al.(2019{\natexlab{a}})Wang, Liu, Zhong, Fu, Qiu, and Huang}]{wang2019exploring}
Danqing Wang, Pengfei Liu, Ming Zhong, Jie Fu, Xipeng Qiu, and Xuanjing Huang. 2019{\natexlab{a}}.
\newblock Exploring domain shift in extractive text summarization.
\newblock \emph{arXiv preprint arXiv:1908.11664}.

\bibitem[{Wang et~al.(2019{\natexlab{b}})Wang, Liu, Zhong, Fu, Qiu, and Huang}]{DBLP:journals/corr/abs-1908-11664}
Danqing Wang, Pengfei Liu, Ming Zhong, Jie Fu, Xipeng Qiu, and Xuanjing Huang. 2019{\natexlab{b}}.
\newblock \href {http://arxiv.org/abs/1908.11664} {Exploring domain shift in extractive text summarization}.
\newblock \emph{CoRR}, abs/1908.11664.

\bibitem[{Wolf et~al.(2020)Wolf, Debut, Sanh, Chaumond, Delangue, Moi, Cistac, Rault, Louf, Funtowicz, Davison, Shleifer, von Platen, Ma, Jernite, Plu, Xu, Le~Scao, Gugger, Drame, Lhoest, and Rush}]{wolf-etal-2020-transformers}
Thomas Wolf, Lysandre Debut, Victor Sanh, Julien Chaumond, Clement Delangue, Anthony Moi, Pierric Cistac, Tim Rault, Remi Louf, Morgan Funtowicz, Joe Davison, Sam Shleifer, Patrick von Platen, Clara Ma, Yacine Jernite, Julien Plu, Canwen Xu, Teven Le~Scao, Sylvain Gugger, Mariama Drame, Quentin Lhoest, and Alexander Rush. 2020.
\newblock \href {https://doi.org/10.18653/v1/2020.emnlp-demos.6} {Transformers: State-of-the-art natural language processing}.
\newblock In \emph{Proceedings of the 2020 Conference on Empirical Methods in Natural Language Processing: System Demonstrations}, pages 38--45, Online. Association for Computational Linguistics.

\bibitem[{Wu et~al.(2021)Wu, Li, Xiao, Liu, Cao, Li, Wu, and Wang}]{wu-etal-2021-bass}
Wenhao Wu, Wei Li, Xinyan Xiao, Jiachen Liu, Ziqiang Cao, Sujian Li, Hua Wu, and Haifeng Wang. 2021.
\newblock \href {https://doi.org/10.18653/v1/2021.acl-long.472} {{BASS}: Boosting abstractive summarization with unified semantic graph}.
\newblock In \emph{Proceedings of the 59th Annual Meeting of the Association for Computational Linguistics and the 11th International Joint Conference on Natural Language Processing (Volume 1: Long Papers)}, pages 6052--6067, Online. Association for Computational Linguistics.

\bibitem[{Xiao and Carenini(2022)}]{entity}
Wen Xiao and Giuseppe Carenini. 2022.
\newblock \href {https://doi.org/10.48550/arXiv.2209.03479} {Entity-based spancopy for abstractive summarization to improve the factual consistency}.
\newblock \emph{CoRR}, abs/2209.03479.

\bibitem[{Xu et~al.(2020)Xu, Li, Yuan, Wu, He, and Zhou}]{SAGCopy}
Song Xu, Haoran Li, Peng Yuan, Youzheng Wu, Xiaodong He, and Bowen Zhou. 2020.
\newblock \href {https://doi.org/10.18653/v1/2020.acl-main.125} {Self-attention guided copy mechanism for abstractive summarization}.
\newblock In \emph{Proceedings of the 58th Annual Meeting of the Association for Computational Linguistics, {ACL} 2020, Online, July 5-10, 2020}, pages 1355--1362. Association for Computational Linguistics.

\bibitem[{Yang et~al.(2020{\natexlab{a}})Yang, Zhu, Gmyr, Zeng, Huang, and Darve}]{ted}
Ziyi Yang, Chenguang Zhu, Robert Gmyr, Michael Zeng, Xuedong Huang, and Eric Darve. 2020{\natexlab{a}}.
\newblock \href {https://doi.org/10.18653/v1/2020.findings-emnlp.168} {{TED:} {A} pretrained unsupervised summarization model with theme modeling and denoising}.
\newblock In \emph{Findings of the Association for Computational Linguistics: {EMNLP} 2020, Online Event, 16-20 November 2020}, volume {EMNLP} 2020 of \emph{Findings of {ACL}}, pages 1865--1874. Association for Computational Linguistics.

\bibitem[{Yang et~al.(2020{\natexlab{b}})Yang, Zhu, Gmyr, Zeng, Huang, and Darve}]{yang2020ted}
Ziyi Yang, Chenguang Zhu, Robert Gmyr, Michael Zeng, Xuedong Huang, and Eric Darve. 2020{\natexlab{b}}.
\newblock Ted: A pretrained unsupervised summarization model with theme modeling and denoising.
\newblock In \emph{Findings of the Association for Computational Linguistics: EMNLP 2020}, pages 1865--1874.

\bibitem[{Yu et~al.(2021)Yu, Liu, and Fung}]{yu-etal-2021-adaptsum}
Tiezheng Yu, Zihan Liu, and Pascale Fung. 2021.
\newblock \href {https://doi.org/10.18653/v1/2021.naacl-main.471} {{A}dapt{S}um: Towards low-resource domain adaptation for abstractive summarization}.
\newblock In \emph{Proceedings of the 2021 Conference of the North American Chapter of the Association for Computational Linguistics: Human Language Technologies}, pages 5892--5904, Online. Association for Computational Linguistics.

\bibitem[{Zhang et~al.(2022)Zhang, Yavuz, Kry{\'s}ci{\'n}ski, Hashimoto, and Zhou}]{zhang2022improving}
Haopeng Zhang, Semih Yavuz, Wojciech Kry{\'s}ci{\'n}ski, Kazuma Hashimoto, and Yingbo Zhou. 2022.
\newblock Improving the faithfulness of abstractive summarization via entity coverage control.
\newblock In \emph{Findings of the Association for Computational Linguistics: NAACL 2022}, pages 528--535.

\bibitem[{Zhang et~al.(2020{\natexlab{a}})Zhang, Zhao, Saleh, and Liu}]{pegasus}
Jingqing Zhang, Yao Zhao, Mohammad Saleh, and Peter~J. Liu. 2020{\natexlab{a}}.
\newblock \href {http://proceedings.mlr.press/v119/zhang20ae.html} {{PEGASUS:} pre-training with extracted gap-sentences for abstractive summarization}.
\newblock In \emph{Proceedings of the 37th International Conference on Machine Learning, {ICML} 2020, 13-18 July 2020, Virtual Event}, volume 119 of \emph{Proceedings of Machine Learning Research}, pages 11328--11339. {PMLR}.

\bibitem[{Zhang et~al.(2020{\natexlab{b}})Zhang, Zhao, Saleh, and Liu}]{DBLP:conf/icml/ZhangZSL20}
Jingqing Zhang, Yao Zhao, Mohammad Saleh, and Peter~J. Liu. 2020{\natexlab{b}}.
\newblock \href {http://proceedings.mlr.press/v119/zhang20ae.html} {{PEGASUS:} pre-training with extracted gap-sentences for abstractive summarization}.
\newblock In \emph{Proceedings of the 37th International Conference on Machine Learning, {ICML} 2020, 13-18 July 2020, Virtual Event}, volume 119 of \emph{Proceedings of Machine Learning Research}, pages 11328--11339. {PMLR}.

\bibitem[{Zhang et~al.(2018)Zhang, Dinan, Urbanek, Szlam, Kiela, and Weston}]{DBLP:conf/acl/KielaWZDUS18}
Saizheng Zhang, Emily Dinan, Jack Urbanek, Arthur Szlam, Douwe Kiela, and Jason Weston. 2018.
\newblock \href {https://doi.org/10.18653/v1/P18-1205} {Personalizing dialogue agents: {I} have a dog, do you have pets too?}
\newblock In \emph{Proceedings of the 56th Annual Meeting of the Association for Computational Linguistics, {ACL} 2018, Melbourne, Australia, July 15-20, 2018, Volume 1: Long Papers}, pages 2204--2213. Association for Computational Linguistics.

\bibitem[{Zhang et~al.(2020{\natexlab{c}})Zhang, Kishore, Wu, Weinberger, and Artzi}]{bertscore}
Tianyi Zhang, Varsha Kishore, Felix Wu, Kilian~Q. Weinberger, and Yoav Artzi. 2020{\natexlab{c}}.
\newblock \href {https://openreview.net/forum?id=SkeHuCVFDr} {Bertscore: Evaluating text generation with {BERT}}.
\newblock In \emph{8th International Conference on Learning Representations, {ICLR} 2020, Addis Ababa, Ethiopia, April 26-30, 2020}. OpenReview.net.

\bibitem[{Zhao et~al.(2022)Zhao, Zheng, Zeng, He, Xu, Jiang, Wu, and Wu}]{zhao-etal-2022-domain}
Lulu Zhao, Fujia Zheng, Weihao Zeng, Keqing He, Weiran Xu, Huixing Jiang, Wei Wu, and Yanan Wu. 2022.
\newblock \href {https://doi.org/10.18653/v1/2022.naacl-main.357} {Domain-oriented prefix-tuning: Towards efficient and generalizable fine-tuning for zero-shot dialogue summarization}.
\newblock In \emph{Proceedings of the 2022 Conference of the North American Chapter of the Association for Computational Linguistics: Human Language Technologies}, pages 4848--4862, Seattle, United States. Association for Computational Linguistics.

\bibitem[{Zhou et~al.(2017)Zhou, Yang, Wei, and Zhou}]{zhou-etal-2017-selective}
Qingyu Zhou, Nan Yang, Furu Wei, and Ming Zhou. 2017.
\newblock \href {https://doi.org/10.18653/v1/P17-1101} {Selective encoding for abstractive sentence summarization}.
\newblock In \emph{Proceedings of the 55th Annual Meeting of the Association for Computational Linguistics (Volume 1: Long Papers)}, pages 1095--1104, Vancouver, Canada. Association for Computational Linguistics.

\bibitem[{Zhou et~al.(2018)Zhou, Yang, Wei, and Zhou}]{DBLP:conf/aaai/ZhouYWZ18}
Qingyu Zhou, Nan Yang, Furu Wei, and Ming Zhou. 2018.
\newblock \href {https://www.aaai.org/ocs/index.php/AAAI/AAAI18/paper/view/16323} {Sequential copying networks}.
\newblock In \emph{Proceedings of the Thirty-Second {AAAI} Conference on Artificial Intelligence, (AAAI-18), the 30th innovative Applications of Artificial Intelligence (IAAI-18), and the 8th {AAAI} Symposium on Educational Advances in Artificial Intelligence (EAAI-18), New Orleans, Louisiana, USA, February 2-7, 2018}, pages 4987--4995. {AAAI} Press.

\end{thebibliography}
\newpage
\appendix
\section{Example}
\label{appa}
Table \ref{tab:eg4indicatorapp} shows the example in Table \ref{tab:eg4indicator} in details where only irrelevant sentences are omitted. We also present the overlapped bi-grams that are used for $C$ in Eq. \ref{copylabel} and $\mathcal{L}_{copy}$ in Eq. \ref{indicator}.

The example shows the nature of the copied bi-grams to some extent. (i) Intersection with entities is usually contained, e.g.,  Hollywood actor, WikiLeaks founder, Ecuadorian Embassy. (ii) Some idiomatic expressions are contained, e.g., take shelter in, to do something, by doing something.

\begin{table}[htbp]
	\centering\small
	\begin{tabularx}{\linewidth}{X}
		\toprule
		\textbf{Article:}   \\
		 Hollywood actor John Cusack is the latest supporter to visit WikiLeaks founder Julian Assange in his continued stay at the Ecuadorian Embassy... Assange has avoided being extradited to Sweden by taking shelter in the Ecuadorean Embassy in London since 2012... Hollywood actor John Cusack (pictured right) is the latest supporter to visit WikiLeaks founder Julian Assange (left) in his continued stay at the Ecuadorian Embassy - where he has remained since 2012... The Australian has been granted political asylum by Ecuador but faces arrest if he leaves the embassy... Mr Assange believes if he is sent to Sweden he will be extradited to the US, where he could face 35 years in prison for publishing on WikiLeaks classified documents related to US activities in Iraq and Afghanistan...
		\\
		\cdashline{1-1}[4pt/2pt]
		\textbf{Summary:} \\
            \textcolor{blue}{Hollywood actor} is \textcolor{blue}{latest supporter to visit WikiLeaks founder} Assange. Pictured arriving \textcolor{blue}{at the Ecuadorian Embassy} where Assange is staying. Assange is avoiding extradition \textcolor{blue}{to Sweden by taking shelter in} embassy. \\
            \cdashline{1-1}[4pt/2pt]
		\textbf{bi-grams} \\
            (Hollywood, actor), (latest, supporter), (supporter, to), (to, visit), (visit, WikiLeaks), (WikiLeaks, founder), (at, the), (the, Ecuador), (Ecuador, an), (an, Embassy), (to, Sweden), (Sweden, by), (by, taking), (taking, shelter), (shelter, in) \\
		\bottomrule
	\end{tabularx}
	\caption{An example for the copying indicator from CNN/DM.}
        \label{tab:eg4indicatorapp}
\end{table}%


\section{Experimental Details}\label{appb}
In empirical studies, our implementation is following \textit{Transformers} library. 
The best scores are selected by the validation set and reported in the tables.
All experiments are conducted on 32GB NVIDIA V100 GPUs.
Our best setup of fine-tuning on CNN/DM is 64 for batch size, 5e-5 for learning rate, 30000 for step numbers.
For pre-training, the better setup is 1e-6 for learning rate, 80 for batch size.
In pre-training data construction, the parameters we use are as follows: the maximum sentence number for $D_{chunk}$ is 8, the minimum is 4, $min_{EFD}$ is 3, $m\%$ is 0.25, $\hat{m}$ is 3.

Since the implementations of ROUGE-L may differ between packages, we clarify that ROUGE-L scores that we report in our paper are from \textit{rouge-scorer} package, which split texts using $\backslash n$. To avoid misunderstanding or confusion, we present another ROUGE-L that ignores $\backslash n$ in Table \ref{rougel}.
\begin{table}[htb]
	\centering\small
	{\begin{tabular}{p{3cm}p{0.6cm}p{0.6cm}p{0.6cm}}
		\toprule
            \textbf{Model} & \textbf{R$_{1}$} &\textbf{R$_{2}$} &\textbf{R$_{L}$}\\ 
		\midrule
            PROM & 44.47 & 21.59 & 31.06 \\
            PROM$_{two\_stage}$ & 44.35 & 21.61 & 31.15\\
            PROM$_{pre-train}$ & 44.59 & 21.66 & 31.37 \ \\
            \midrule
            \textbf{Dataset} & \textbf{R$_{1}$} &\textbf{R$_{2}$} &\textbf{R$_{L}$}\\ 
		\midrule
            CNN/DM & 37.87 &15.91 &  24.93 \\
            XSum & 22.96 & 6.05 & 17.27 \\
            NYT & 36.95 & 17.21 &  25.67\\
            WikiHow & 25.90 & 6.37 &  16.17 \\
            BillSum & 44.35 & 21.61 &  23.64 \\
            arXiv & 40.05 &14.66 &  19.93 \\
		\bottomrule
	\end{tabular}
	}
        \caption{ROUGE $\operatorname{F}_{1}$ scores. The upper part is fine-tuning results, and the lower part is zero-shot results.}
	\label{rougel}
\end{table}

Before the pre-training, a pilot experiment is conducted to compare PROM with the baseline model BART. They are trained on a mini pre-training dataset derived from a small subset of the corpora. Scores are shown in Table \ref{pilot}.
\begin{table}[tbh]
	\centering\small
	{\begin{tabular}{p{1.7cm}p{0.4cm}p{0.4cm}p{0.5cm}p{0.4cm}p{0.4cm}p{0.4cm}}
		\toprule
		{\textbf{Model}} & \makebox[0.4cm][c]{\textbf{R$_{1}$}} & \textbf{R$_{2}$} & \textbf{R$_{L}$} & \textbf{R$_{1}$} & \textbf{R$_{2}$} & \textbf{R$_{L}$} \\ 
            \midrule
            & \multicolumn{3}{c}{\textbf{CNN/DM}} &\multicolumn{3}{c}{\textbf{BillSum}}\\
		\midrule
            BART  & \makebox[0.4cm][c]{33.47} &\makebox[0.4cm][c]{12.90} &\makebox[0.4cm][c]{30.59}   &\makebox[0.4cm][c]{36.97} & \makebox[0.4cm][c]{13.10} & \makebox[0.4cm][c]{30.93} \\ 
            
            PROM$_\textup{mini}$  & \makebox[0.4cm][c]{34.34} &\makebox[0.4cm][c]{13.49} &\makebox[0.4cm][c]{31.32}  &\makebox[0.4cm][c]{37.56} & \makebox[0.4cm][c]{12.97} & \makebox[0.4cm][c]{31.53} \\
            
		\midrule
            \multirow{2}{*}{}  &\multicolumn{3}{c}{\textbf{arXiv}} \\ 
            \midrule
            BART  &  \makebox[0.4cm][c]{33.25} & \makebox[0.4cm][c]{9.8} & \makebox[0.4cm][c]{28.82}  &&\\ 
            
            PROM$_\textup{mini}$   & \makebox[0.4cm][c]{34.23} & \makebox[0.4cm][c]{9.69} & \makebox[0.4cm][c]{29.84} && \\
            \bottomrule
	\end{tabular}
	}
        \caption{Zero-shot ROUGE $\operatorname{F}_1$ numbers of pilot experiments.}
	\label{pilot}
\end{table}

\section{Human Evaluation}
\label{appc}
We present our implementation of human evaluation in this section. We first show the instruction that explains the experiment object and measurement, i.e., faithfulness, informativeness, and readability.
Then, we show the scoring rules.

\begin{prompt}[title={Instructions}]
We illustrate the features: faithfulness, informativeness, and readability.

1) \textbf{Faithfulness.} 
Faithfulness means factual consistency with the context. Please avoid using general knowledge, and only consider it in the context of the provided document. The summary is inconsistent if facts in the summary are not supported by the document. Two typical cases are conflict and hallucination.

\quad (\romannumeral1) The summary contradicts the information in the document. The summary might say "A fire broke out in Seattle", but the document says it broke out in Portland. Or the summary might say "the Republicans won the election", but the document indicates the Democrats won instead.

\quad (\romannumeral2) The summary adds (hallucinates) a fact that is not mentioned anywhere in the document. For example, the summary might say that "A fire broke out at 2 am", but the document does not mention the time when the fire broke out.

2) \textbf{Informativeness.} It means that a summary expresses the main points of the document. A summary should contain relevant and important information and few unimportant details. If you select the summary to be not consistent with the document, please only consider the consistent information when evaluating this category.

3) \textbf{Readability.} The summaries are written by human or generated by language models. A summary is readable/fluent if free from language problems. A less readable summary is confusing and difficult to understand.

We present an example in Table \ref{tab:addlabel}. This article reports the accidental death of Alexys Brown. The main information is the accident and the appeal to raise money and minor points can be the investigation, post-mortem examination, etc. 
Summary 1 can be reasonable and acceptable.
Summary 2 misses a major point, the appeal, thus not informative. 
Both summary 3 and summary 4 show unfaithfulness. 
Summary 3 makes a factual mistake that Alexys died \textit{of cancer}. This contradicts the article. Summary 4 adds a hallucination that Alexys is \textit{three-year-old}. 
Summary 5 is confusing because grammar flaws impair readability.
\end{prompt}
\begin{prompt}[title={Scoring rules}]
Please annotate which one of the 2 summaries is better in the four aspects separately. For example, if summary \#1 is better than summary \#2 in the aspect of informativeness, then type “1” behind `*****Informativeness:`. It means summary \#1 wins over summary \#2 on informativeness. And if summary \#2 wins in readability, then type “2” behind `*****Readability:`. If the two summaries draw with each other (come out even) in an aspect, then type “0” in the cell below that aspect. 
There may be repeated summaries of an article, and please make sure that their scores are all 0.

Your results are 3 integers for each sample. The scores are in the order of \textit{faithfulness, informativeness, readability}. The format is shown in Table \ref{tab:format}.
\end{prompt}
\begin{table*}[htbp]
	\centering
	\begin{tabularx}{\textwidth}{X}
		\toprule
		\textbf{Article:}   \\
		 Alexys Brown, also known as Lexi, died at her home in Emmadale Close, Weymouth, on Thursday. An investigation is under way to discover how she became trapped. A post-mortem examination is due to be carried out this week. It was originally hoped the appeal would raise £2,000. Alison Record, who started the Just Giving appeal, said she was "heart broken" over the death. “Everybody by now has heard of the terrible tragedy the Brown family have suffered with the loss of their beautiful and beloved little girl Lexi,” the appeal page reads. Many other comments have been posted on the appeal page. Steph Harris said: “Thinking of you all at this devastating time, fly high beautiful princess. Love Steph and family xxx” Lesley Andrews added: “No amount of money will take away the pain, but so much love comes with every penny. Take care. xx” Aster Group, the housing association responsible for managing the home, is assisting with the police investigation. The Health and Safety Executive (HSE) is also investigating. Dorset County Council said it had not installed the disabled lift at the property.
		\\
            \midrule
            \textbf{Summary \#1:}\\
            An appeal to raise money for the family of a girl who died after getting stuck in a lift was originally hoped for raising 2,000 pounds. \\
		\midrule
		\textbf{Summary \#2 (informativeness):} \\
            Alexys Brown, also known as Lexi, died at her home in Emmadale Close, Weymouth, on Thursday. \\
		\midrule
            \textbf{Summary \#3 (faithfulness):} \\
            Alexys Brown, also known as Lexi, died \underline{of cancer}. The appeal was originally hoped for raising 2,000 pounds. \\
            \midrule
            \textbf{Summary \#4 (faithfulness):} \\
            An appeal to raise money for Alexys Brown, a \underline{three-year-old girl} who died after getting stuck in a lift was originally hoped for raising 2,000 pounds.\\
            \midrule
            \textbf{Summary \#5 (readability):} \\
             An appeal to raise the family of Alexys Brown became trapped in a lift would raise 2,000 pounds.\\
		\bottomrule
	\end{tabularx}%
	\caption{An example for the instructions.}
 \label{tab:addlabel}
\end{table*}

\begin{table*}[htbp]
	\centering
	\begin{tabularx}{\textwidth}{X}
		\toprule
		\textbf{Article:}   \\
		 Alexys Brown, also known as Lexi, died at her home in Emmadale Close, Weymouth, on Thursday. An investigation is under way to discover how she became trapped. A post-mortem examination is due to be carried out this week. It was originally hoped the appeal would raise £2,000. Alison Record, who started the Just Giving appeal, said she was "heart broken" over the death. “Everybody by now has heard of the terrible tragedy the Brown family have suffered with the loss of their beautiful and beloved little girl Lexi,” the appeal page reads. Many other comments have been posted on the appeal page. Steph Harris said: “Thinking of you all at this devastating time, fly high beautiful princess. Love Steph and family xxx” Lesley Andrews added: “No amount of money will take away the pain, but so much love comes with every penny. Take care. xx” Aster Group, the housing association responsible for managing the home, is assisting with the police investigation. The Health and Safety Executive (HSE) is also investigating. Dorset County Council said it had not installed the disabled lift at the property.
		\\
            \midrule
            \textbf{Summary \#1: }\\
             Alexys Brown, also known as Lexi, died at her home in Emmadale Close, Weymouth, on Thursday. \\
             \textbf{Summary \#2: }\\
             Alexys Brown, also known as Lexi, died \underline{of cancer}. The appeal was originally hoped for raising 2,000 pounds. \\
            \midrule
            \textbf{Scores: }\\
            *****Faithfulness: 1\\
            *****Informativeness: 2\\
            *****Readability: 0\\
		\bottomrule
	\end{tabularx}%
	\caption{An example for the scoring format.}
 \label{tab:format}
\end{table*}

\end{document}